\documentclass[letterpaper,twocolumn,10pt]{article}
\usepackage{usenix2019_v3}
\usepackage{tikz}
\usepackage{amsmath}
\usepackage{filecontents}
\usepackage{graphicx} 
\usepackage{mathtools,physics}
\usepackage[most]{tcolorbox}
\usepackage{appendix}

\usepackage{cite}
\usepackage{mathtools}
\usepackage{amssymb}
\usepackage{multirow}
\usepackage{float}
\usepackage{comment}
\usepackage{booktabs}
\usepackage{hyperref}
\usepackage{diagbox}

\usepackage{multirow}
\usepackage{array}
\usepackage{threeparttable}
\usepackage{booktabs}

\usepackage{arydshln}
\usepackage{graphicx}

\usepackage{subcaption}

\begin{document}

\title{Efficiency Robustness of Dynamic Deep Learning Systems}

\author{
{\rm Ravishka Rathnasuriya}, {\rm Tingxi Li$^*$}, {\rm Zexin Xu$^*$}, {\rm Zihe Song$^*$}, {\rm Mirazul Haque}, {\rm Simin Chen}, {\rm Wei Yang}
\\The University of Texas at Dallas\\
\{ravishka.rathnasuriya, tingxi.li, zexin.xu, zihe.song, mirazul.haque, simin.chen, wei.yang\}@utdallas.edu}


\maketitle
\def\thefootnote{*}\footnotetext{These authors contributed equally as second authors.}\def\thefootnote{\arabic{footnote}}

\begin{abstract}

Deep Learning Systems (DLSs) are increasingly deployed in real-time applications, including those in resource-constrained environments such as mobile and IoT devices. To address efficiency challenges, Dynamic Deep Learning Systems (DDLSs) adapt inference computation based on input complexity, reducing overhead. While this dynamic behavior improves efficiency, such behavior introduces new attack surfaces. In particular, \textbf{efficiency adversarial attacks} exploit these dynamic mechanisms to degrade system performance.

This paper systematically explores efficiency robustness of DDLSs, presenting the first comprehensive taxonomy of efficiency attacks. We categorize these attacks based on three dynamic behaviors: (i) attacks on dynamic computations per inference, (ii) attacks on dynamic inference iterations, and (iii) attacks on dynamic output production for downstream tasks. Through an in-depth evaluation, we analyze adversarial strategies that target DDLSs efficiency and identify key challenges in securing these systems. In addition, we investigate existing defense mechanisms, demonstrating their limitations against increasingly popular efficiency attacks and the necessity for novel mitigation strategies to secure future adaptive DDLSs.  

\end{abstract}

\section{Introduction}

Deep Learning Systems (DLSs) have demonstrated exceptional performance across diverse domains, including object recognition, speech processing, and machine translation~\cite{han2021dynamic}. The adoption of these models in real-time applications, such as mobile platforms, unmanned aerial vehicles, and edge computing devices, underscores the critical need for computational efficiency. Despite their success, the resource-intensive nature of deep learning inference presents considerable challenges, particularly when deployed on power-constrained hardware.




To address these limitations, Dynamic Deep Learning Systems (DDLSs)~\cite{skipnet,conditional,lowrank,runtimeprune,adaptive} have been developed. Unlike traditional static models~\cite{googlenet15cvpr,resnet16cvpr,mobilenet}, DDLSs adapt computational workloads based on input complexity, optimizing performance while reducing latency and energy consumption. Examples include multi-exit architectures\textcolor{blue}{~\cite{u2022unsupervisedearlyexitdnns,salehi2023sharcsefficienttransformersrouting}}, token-pruned transformers\textcolor{blue}{~\cite{gao2024energylatencymanipulationmultimodallarge, NICGSlowDown}}, and dynamic neural networks that adjust processing depth\textcolor{blue}{~\cite{skipnet, conditional, lowrank, runtimeprune, adaptive}} or activation sparsity\textcolor{blue}{~\cite{sparsity, 2.5baischer2021learning, hardwareDNN1, hoefler2021sparsity}}.  A prominent example of dynamic computation is found in models like GPT~\cite{openai2023gpt4}, where computational complexity varies depending on the input and the progression of the model's output. However, such an adaptability also introduces new security vulnerabilities. 

Recent studies have shown that adversaries can exploit these dynamic mechanisms to launch efficiency adversarial attacks~\cite{ILFO,chen2022deepperform,EREBA,sponge}, where carefully crafted inputs induce excessive computational cost.  Unlike traditional adversarial attacks, which focus on misclassification by perturbing inputs, efficiency attacks manipulate inference-time computational costs without necessarily affecting output correctness. These attacks degrade system efficiency, leading to increased latency, increased energy consumption, and, in extreme cases, denial-of-service conditions. The consequences extend beyond inefficiency: they can disrupt mission-critical applications, rapidly drain mobile device batteries, and escalate operational costs in cloud environments.
For example, similar to resource exhaustion denial-of-service attacks~\cite{regdos} in traditional computing, an efficiency attack can force an AI-powered security camera to perform excessive computations, delaying threat detection and rendering the system ineffective in real-time scenarios. In real-time assistive vision systems, adversarial inputs can delay object recognition, compromising user safety. In cloud-based AI services, attackers can generate queries that inflate computational costs, leading to resource exhaustion and financial strain on service providers. 

Although various efficiency attack strategies\cite{NICGSlowDown,NMTSloth,sponge,ILFO,GradAuto, chen2022deepperform,hong2020slowdownattack,GradMDM, PhantomSponge, liu2023slowlidar,NODEA,malice,EREBA, efficfrog, sparsity, cina2022spongepoisoning} have been identified across different DDLS architectures~\cite{skipnet,chen2018neural,SNN1,SNN2,SNN3,2.4.1anderson2018bottom, 2.4.1cornia2020meshed, 2.4.1del2020ratt,2.4.1gan2017semantic, 2.4.1pan2020x, 2.4.1radford2018improving, 2.4.1vaswani2017attention,sparsity, 2.5baischer2021learning, hardwareDNN1}, there is a lack of a unified framework that categorizes and analyzes these threats holistically. Without a structured understanding, the AI community lacks systematic defense strategies, leaving DDLSs highly vulnerable to evolving efficiency attacks. Such a gap threatens the reliability of AI-powered systems in critical applications, from cybersecurity to autonomous systems.

To address the aforementioned challenge, this paper provides the first comprehensive systematization of efficiency adversarial attacks on DDLSs. We systematically analyze these threats, offering a structured framework that enhances the robustness of DDLSs against such vulnerabilities.

Our key contributions include:
\begin{itemize}
    \item \textbf{Problem Formulation \& Taxonomy:} We define efficiency attacks (Section~\ref{problem}) and establish a comprehensive taxonomy of these attacks on DDLSs (Section~\ref{sec:tax}), including an exploration of dynamic behaviors (Section ~\ref{behavior}) that lead to such vulnerabilities.
    \item \textbf{Attack Mechanisms \& Defense Strategies}: We analyze the mechanisms behind efficiency attacks (Section ~\ref{attacks}) and assess existing defense strategies (Section~\ref{defense}). Our empirical evaluation (Section ~\ref{sec:def}) highlights that beyond detecting adversarial examples, defenses must process them efficiently to enhance robustness.
    \item \textbf{State-of-the-Art Analysis \& Research Directions}: We provide a detailed analysis of current research, identifying key challenges and proposing future research directions in the domain of DDLS security.
    
\end{itemize}

This Systematization of Knowledge benefits both deep learning researchers and practitioners by providing a formal foundation for understanding efficiency attacks, examining their real-world impact, and outlining strategies for building more resilient and efficient DDLSs.

\section{Understanding Efficiency Robustness of DDLSs}
\label{background}

\begin{figure*}[t]
    \centering
    
    \includegraphics[width=\linewidth]{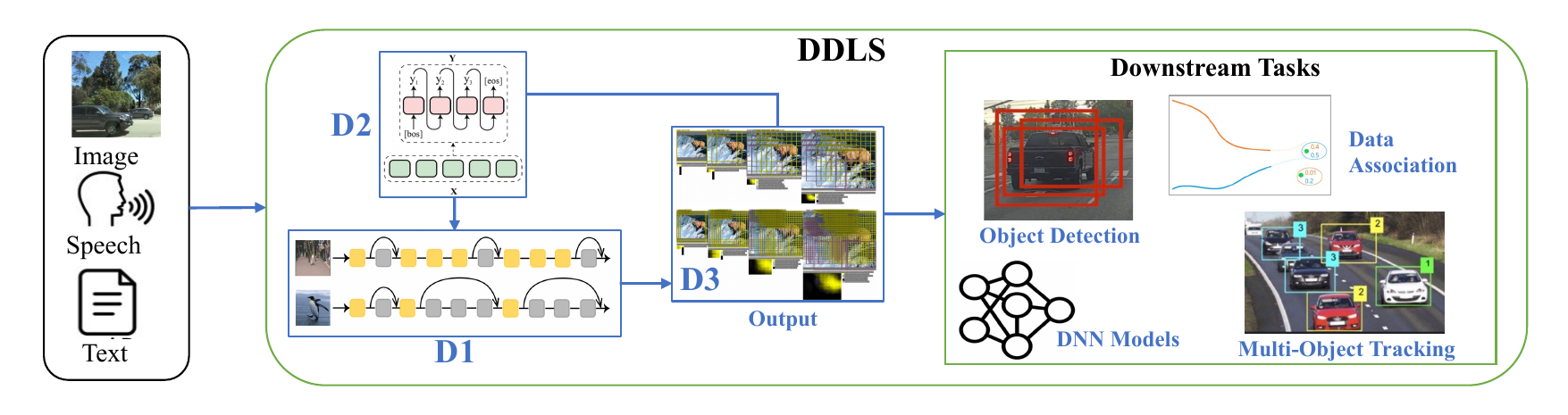}
   
    \caption{Illustration of Dynamic Behavior (\textbf{\texttt{D}}) of DDLS. \textbf{\texttt{D1}} examines the behavior where computational outputs fluctuate across individual inference iterations. \textbf{\texttt{D2}} focuses on attacks that alter the number of iterations needed to complete inference. \textbf{\texttt{D3}} involves attacks that escalate the number of generated outputs, thereby increasing the overall computational load.}
    \label{fig:ddls}
\end{figure*}



This section provides a structured overview of efficiency attacks on DDLSs, defining their scope, the key dynamic behaviors intrinsic to DDLSs, and outlining relevant threat models. Together, these elements form a foundation for analyzing attack strategies.

\subsection{Efficiency Attacks: Definition and Scope}
\label{problem}

Efficiency attacks exploit the dynamic nature of DDLSs to maximize computational cost. Unlike traditional adversarial attacks that induce misclassification, these attacks degrade system performance by inflating resource consumption. 


We define an efficiency attack as a constrained optimization problem. Given a neural network $\mathcal{N}(\cdot)$, an input $x \in \mathcal{X}$, and a hardware accelerator $\mathcal{H}$, the computational cost is represented as $\mathcal{C}(\mathcal{N}, \mathcal{H}, x)$ measured in energy consumption, latency, or floating point operations~(FLOPs). The adversary seeks an optimal perturbation $\delta$ that maximizes computational cost while remaining imperceptible and within the valid input space: 


\begin{equation}
\begin{split}
    &\text{maximize} \quad \mathcal{C}(\mathcal{N}, \mathcal{H}, x + \delta)\   \\ 
& \text{subject to} \quad ||\delta|| \leq \epsilon, \quad x + \delta \in \mathcal{X}
\end{split}
\end{equation}

The objective is to increase computational cost without violating input constraints ($||\delta|| \leq \epsilon$), exposing a fundamental vulnerability in efficiency-optimized DDLSs.

\subsection{Adversarial Exploitation of Dynamic Behaviors of DDLSs}
\label{behavior}


Figure~\ref{fig:ddls} illustrates the dynamic behaviors of DDLSs, where the system adapts computations to minimize resource usage as inputs, such as images, speech, or text, are processed. These behaviors span three dimensions: (\textbf{\texttt{D1}}) varying computation per inference, (\textbf{\texttt{D2}}) varying number of inference iterations per input, and (\textbf{\texttt{D3}}) varying number of inputs for the downstream modules. While these mechanisms improve efficiency, they also expose vulnerabilities that adversaries can exploit to maximize computational cost.


Dynamic computation per inference (\textbf{\texttt{D1}}) describes a behavior that adjust their internal execution paths during a single forward pass. Components such as layers, neurons, or solver steps are selectively activated or skipped based on gating signals or confidence scores. Common mechanisms include \textit{early exits, conditional skipping, adaptive step sizing}, and \textit{activation sparsity}. Adversaries exploit these mechanisms (Section \ref{DNN_arch}) by inducing full-path execution, suppressing early exits or pruning logic, and thereby forcing full-path execution and inflating computation.

Dynamic inference iterations (\textbf{\texttt{D2}}) capture a behavior that generate or refine outputs over iterative inference, with iteration counts determined at runtime.  Such behavior is common in autoregressive and iterative generation models, where outputs (e.g., tokens) are produced step-by-step. Termination is governed by learned halting policies, convergence checks, or stop tokens. Unlike \texttt{D1}’s intra-pass adaptivity, \texttt{D2} reflects iterative adaptivity. Adversaries exploit such a behavior (Section ~\ref{d2}) by crafting inputs that delay termination, thereby inflating computational cost.




Dynamic output production (\textbf{\texttt{D3}}) refers to model behavior where number of outputs varies at inference and influences downstream workload. Such behavior is typical in tasks like object detection or multi-object tracking, where output elements (e.g., bounding boxes) are fed into subsequent modules. Output cardinality depends on internal decisions (e.g., Non-Maximum Suppression~(NMS) thresholds), not fixed formats.
Unlike \texttt{D1} and \texttt{D2}, which affect inference-time cost directly, \texttt{D3} represents post-inference adaptivity, which shifts the burden to downstream stages. Adversaries exploit such behavior (Section~\ref{d3}) by inducing excessive output generation, thereby amplifying system-wide resource consumption.



By targeting these dynamic behaviors, efficiency attacks disrupt the core optimizations of DDLSs, increasing the computational costs.

\subsection{Threat Model}
\label{threat}

An adversary aims to degrade the efficiency of DDLSs by maximizing computational costs during inference time. Through imperceptible adversarial perturbations, the attacker suppresses the model’s dynamic behavior, forcing unnecessary computation without altering output correctness, as discussed in Section~\ref{behavior}.

\noindent\textbf{Adversary's Goal:} 
The adversary seeks to (i) maximize computational cost, degrading system efficiency; (ii) ensure perturbations remain imperceptible to humans; and (iii) maintain real-world plausibility, ensuring adversarial inputs function in practical settings.




\noindent\textbf{Adversary's Knowledge and Capabilities:} 
Efficiency attacks fall into \textbf{evasion} and \textbf{poisoning} attacks, targeting different phases of model execution. 

\textbf{Evasion attacks} occur at inference time, disrupting efficiency. In \textbf{white-box evasion attacks}, the adversary has full access to model architecture, parameters, and intermediate outputs, allowing precise manipulation of inputs to force exhaustive computation by disrupting dynamic mechanisms. In contrast, \textbf{black-box evasion attacks} operate without internal access, relying on observable outputs such as latency or energy consumption to iteratively modify inputs. Using optimization techniques or surrogate models, they infer response patterns to increase computational cost. Black-box attacks often assume that the target model employs a specific type of dynamic mechanism, which influences the attack strategy.

\textbf{Poisoning attacks} target the training phase to embed inefficiencies that persist at inference. These attacks assume that the adversary can inject training data or modify model updates. In data poisoning, adversaries craft inputs to bias model behavior toward increased computational cost. In model poisoning, the training procedure is manipulated to encode inefficient execution paths. Both approaches raise
inference-time costs, undermining dynamic mechanisms designed for efficiency.

\section{Taxonomy}
\label{sec:tax}

\begin{figure*}[t]
    \centering
    
    \includegraphics[width=0.9\textwidth]{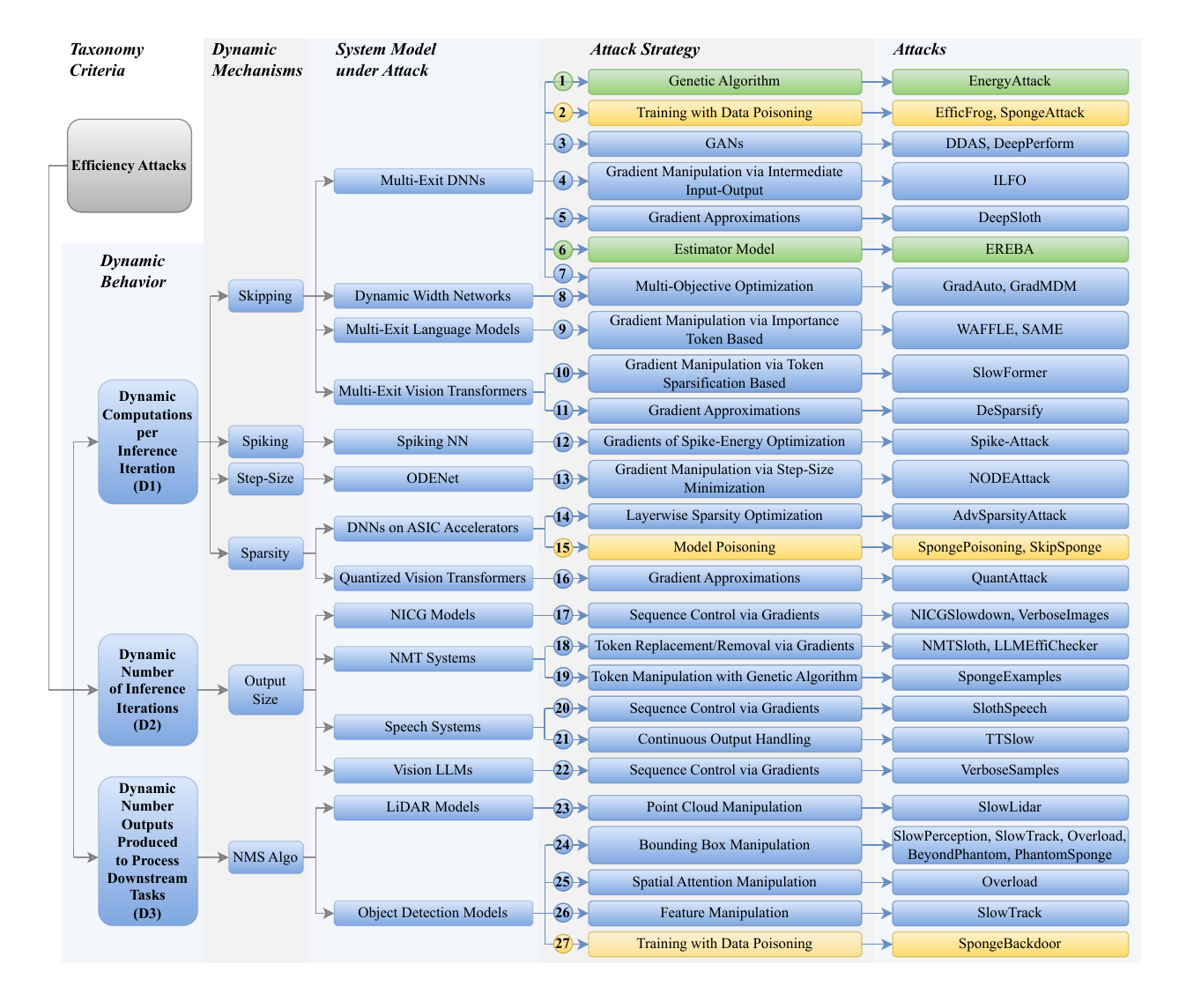}

    \caption{Taxonomy of the efficiency attacks. In attack strategy, green boxes represent black-box evasion attack, blue boxes represent white-box evasion attack, and yellow boxes represent poisoning attack.}
    \label{fig:taxonomy}
\end{figure*}

This section presents a structured taxonomy of efficiency attacks on DDLSs, categorized by the dynamic behaviors they exploit, as illustrated in Figure~\ref{fig:taxonomy}. The taxonomy adopts a hierarchical structure, beginning with three core dynamic behavioral categories: \textbf{\texttt{D1}}, \textbf{\texttt{D2}}, and \textbf{\texttt{D3}}, introduced in Section~\ref{behavior}. Each category reflects a distinct dynamic mechanism of DDLSs that adversaries target to increase computational cost. The taxonomy further maps vulnerable system models to corresponding attack strategies and specific adversarial techniques.

Attacks on \textbf{\texttt{D1}} manipulate computational cost within a single inference iteration. This category includes four major dynamic mechanisms: skipping, spiking, step-size adjustments, and sparsity exploitation. Skipping-based attacks, such as  ILFO~\cite{ILFO}, EREBA~\cite{EREBA}, GradMDM~\cite{GradMDM}, GradAuto~\cite{GradAuto}, EnergyAttack~\cite{du2025energyattack}, EfficFrog~\cite{efficfrog}, SpongeAttack~\cite{10445460},DDAS~\cite{ayyat2022dynamicDDAS}, DeepSloth~\cite{hong2020slowdownattack} and DeepPerform~\cite{chen2022deepperform}, target multi-exit DNNs (ME-DNNs), disrupting early-exit mechanisms to force unnecessary computation. Attacks like SlowFormer~\cite{navaneet2023slowformeruniversaladversarialpatch} and DeSparsify~\cite{yehezkel2024desparsifyadversarialattacktoken} exploit multi-exit vision transformers (ME-ViTs), while WAFFLE~\cite{coalson2023bertwaffle} and SAME~\cite{chen2023dynamic} target multi-exit language models (ME-LMs), nullifying efficiency gains from early exits (Section~\ref{sec:skipping}). Spiking-based attacks, such as SpikeAttack~\cite{SpikeAttack}, increase neuron activation rates in spiking neural networks (SNNs) (Section~\ref{spike}). Step-size manipulation attacks such as NODEAttack~\cite{NODEA} exploit ODE-based models to increase FLOPs by forcing inefficient step-size adjustments (Section~\ref{step-size}). Sparsity-exploitation attacks including SkipSponge~\cite{lintelo2024skipspongeattackspongeweight}, SpongePoisoning~\cite{cina2022spongepoisoning}, AdvSparsityAttack~\cite{sparsity} and QuantAttack~\cite{baras2023quantattackexploitingdynamicquantization} override sparsity-based optimizations in quantized vision transformers and ASIC-deployed DNNs, increasing computational cost (Section~\ref{sparsity}).

Attacks on \textbf{\texttt{D2}} increase the number of inference iterations required for a model to generate an output. This category primarily targets auto-regressive models, including neural image caption generation (NICG) models, neural machine translation (NMT) systems, vision language models (VLMs) , and speech systems (Section~\ref{d2}). NICGSlowdown~\cite{NICGSlowDown} and VerboseImages~\cite{gao2024inducing} exploit the sequential nature of NICG models to force excessive token generation per inference step. NMTSloth~\cite{NMTSloth}, LLMEffiChecker~\cite{feng2024llmeffichecker}, and SpongeExamples~\cite{sponge} manipulate sequence termination conditions in NMT systems, extending inference duration. In speech systems, attacks such as SlothSpeech~\cite{slothspeech} and TTSlow~\cite{gao2024ttslowslowtexttospeechefficiency} introduce perturbations that prolong inference loops. VLMs which rely on iterative refinement are vulnerable to attacks like VerboseSamples~\cite{gao2024energylatencymanipulationmultimodallarge} inducing unnecessary computational steps.

Attacks on \textbf{\texttt{D3}} manipulate the number of outputs generated by a model, increasing downstream computational costs. Unlike \textbf{\texttt{D1}} and \textbf{\texttt{D2}} which target internal inference mechanisms, attacks on \textbf{\texttt{D3}}  impact downstream workloads, particularly in object detection and perception models. Attacks such as SlowLiDAR~\cite{liu2023slowlidar}, SlowPerception~\cite{ma2024slowperceptionphysicalworldlatencyattack}, BeyondPhantom~\cite{10.1145/3649403.3656485}, SpongeBackdoor~\cite{10650435}, and PhantomSponge~\cite{PhantomSponge} introduce perturbations that increase the number of bounding boxes, overloading detection pipelines. SlowTrack~\cite{slowtrack} disrupts multi-object tracking causing excessive object associations, while Overload~\cite{overload} manipulates feature representations to induce granular output generation that increases downstream processing costs.



\begin{figure}[b]
    \centering
    
    \includegraphics[width=\columnwidth]{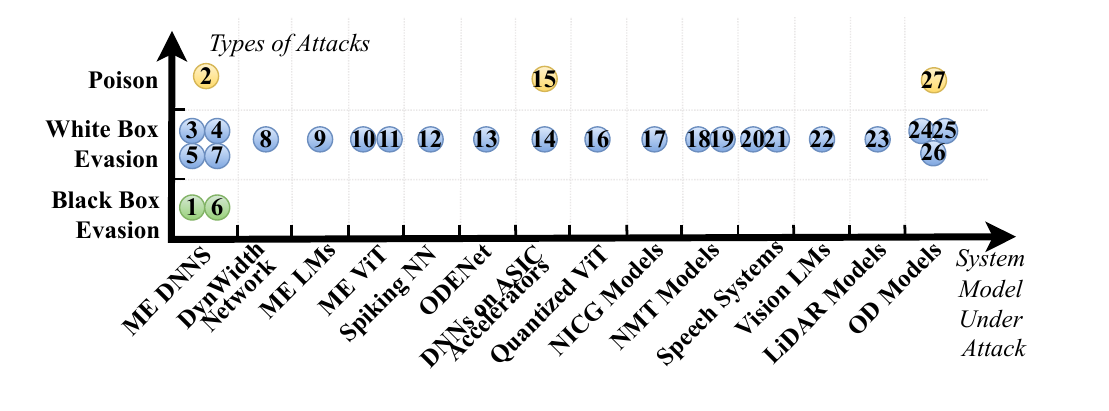}
    
    \caption{Threat models of the efficiency attacks. ME, LM, and OD refer to Multi-Exit, Language Models, and Object Detection, respectively.}
    \label{fig:threat}
\end{figure}

Figure~\ref{fig:threat} illustrates the relationship between attack categories and their corresponding threat models, classifying them as poisoning attacks, white-box evasion attacks, or black-box evasion attacks, as discussed in Section~\ref{threat}. Such mapping establishes a direct link between adversarial capabilities and the computational inefficiencies they induce.


\noindent\textbf{Properties of Efficiency Attacks:} 
Table~\ref{tab:properties_new} summarizes key properties of efficiency attacks, including targeted \textbf{dynamic mechanisms}, \textbf{system models}, \textbf{attack strategies}, \textbf{model access types}, \textbf{attack stages}, \textbf{input modalities}, \textbf{perturbation norms}, and \textbf{attack name}. Each dynamic mechanism is detailed within Table~\ref{tab:properties_new}, specifying whether the attack occurs during training or inference and whether the attack requires white-box or black-box model access. Additionally, the table classifies supported perturbation norms based on the input modality affected. Such a structured analysis provides a foundation for understanding how efficiency attacks systematically degrade DDLS performance. 

\begin{table*}
\begin{center}
\caption{Properties of Efficiency Attacks. The Attack Strategy column corresponds to the ``attack strategies'' presented in Figure~\ref{fig:taxonomy}.} 
\label{tab:properties_new}
\resizebox{\textwidth}{!}{%
\begin{tabular}{c c c c c c c c c c}
\toprule
\multirow{2}{*}{\begin{tabular}[t]{@{}c@{}}\textbf{Dynamic}\\\textbf{Behavior}\end{tabular}} & 
\multirow{2}{*}{\begin{tabular}[t]{@{}c@{}}\textbf{Dynamic}\\\textbf{Mechanism}\end{tabular}} & 
\multirow{2}{*}{\begin{tabular}[t]{@{}c@{}}\textbf{System Model Attacked}\end{tabular}} & 
\multirow{2}{*}{\begin{tabular}[t]{@{}c@{}}\textbf{Attack}\\\textbf{Strategy}\end{tabular}} & 
\multirow{2}{*}{\begin{tabular}[t]{@{}c@{}}\textbf{Model}\\\textbf{Access}\end{tabular}} &
\multirow{2}{*}{\begin{tabular}[t]{@{}c@{}}\textbf{Attack}\\\textbf{Stage}\end{tabular}} &
\multirow{2}{*}{\begin{tabular}[t]{@{}c@{}}\textbf{Efficiency Attack}\\ \textbf{Target}\end{tabular}} &
\multirow{2}{*}{\begin{tabular}[t]{@{}c@{}}\textbf{Input}\\\textbf{Modality}\end{tabular}} &
\multirow{2}{*}{\begin{tabular}[t]{@{}c@{}}\textbf{Perturbation}\\\textbf{Norm}\end{tabular}} &
\multirow{2}{*}{\begin{tabular}[t]{@{}c@{}}\textbf{Attacks}\end{tabular}} \\
\\
\midrule

\multirow{17}{*}{\begin{tabular}[t]{@{}l@{}}Dynamic\\Computations\\per Inference\\Iteration \\(D1)\end{tabular}} 

& \multirow{12}{*}{Skipping} 

& \multirow{4}{*}{\begin{tabular}[t]{@{}c@{}}Multi-Exit DNNs\end{tabular}} & 1 & Black-box & Inference & FLOPs & Image & $L_{2}$, $L_{\infty}$ & EnergyAttack\cite{du2025energyattack} \\
\cdashline{4-10}[0.25pt/1pt]
& & & 2 & White-box & Training & FLOPs & Image & $L_2$ & EfficFrog\cite{efficfrog}, SpongeAttack\cite{10445460} \\
\cdashline{4-10}[0.25pt/1pt]
& & & 3 & White-box & Inference & Latency & Image & $L_2$ & DDAS\cite{ayyat2022dynamicDDAS} \\
\cdashline{4-10}[0.25pt/1pt]
& & & 3 & White-box & Training & FLOPs, Latency & Image & $L_2$, $L_{\infty}$ &  DeepPerform\cite{chen2022deepperform} \\

\cdashline{3-10}[0.25pt/1pt]
& & \multirow{4}{*}{Dynamic Width Networks} & 4 & White-box & Inference & FLOPs, Energy & Image & $L_2$ & ILFO\cite{ILFO} \\
\cdashline{4-10}[0.25pt/1pt]
& & & 5 & White-box & Inference & FLOPs, Latency & Image & $L_1$, $L_2$, $L_{\infty}$ & DeepSloth\cite{hong2020slowdownattack} \\
\cdashline{4-10}[0.25pt/1pt]
& & & 6 & Black-box & Inference & Energy & Image & $L_2$ & EREBA\cite{EREBA} \\
\cdashline{4-10}[0.25pt/1pt]
& & & 7 & White-box & Inference & FLOPs, Energy & Image & $L_2$ & GradAuto\cite{GradAuto}, GradMDM\cite{GradMDM} \\
\cdashline{3-10}[0.25pt/1pt]

& & \multirow{2}{*}{Multi-Exit Language Models} & 8 & White-box & Inference & FLOPs, Energy & Image & $L_2$ & GradAuto\cite{GradAuto}, GradMDM\cite{GradMDM} \\
\cdashline{4-10}[0.25pt/1pt]
& & & 9 & White-box & Inference & FLOPs, Latency & Text & Word Level & WAFFLE\cite{coalson2023bertwaffle}, SAME\cite{chen2023dynamic} \\
\cdashline{3-10}[0.25pt/1pt]

& & \multirow{2}{*}{Multi-Exit Vision Transformers} & 10 & White-box & Inference & FLOPs, Energy & Image & $L_0$ & SlowFormer\cite{navaneet2023slowformeruniversaladversarialpatch} \\
\cdashline{4-10}[0.25pt/1pt]
& & & 11 & White-box & Inference & FLOPs, Latency, Energy & Image & $L_{\infty}$ & DeSparsity\cite{yehezkel2024desparsifyadversarialattacktoken} \\
\cdashline{2-10}[0.25pt/1pt]

& \multirow{1}{*}{Spiking} & \multirow{1}{*}{Spiking NN} & 12 & White-box & Inference & Latency, Energy & Image & $L_{\infty}$ & Spike-Attack\cite{SpikeAttack} \\
\cdashline{2-10}[0.25pt/1pt]

& \multirow{1}{*}{Step-Size} & \multirow{1}{*}{ODENet} & 13 & White-box & Inference & Energy & Image & $L_2$ & NODEAttack\cite{NODEA} \\
\cdashline{2-10}[0.25pt/1pt]

& \multirow{3}{*}{Sparsity} 
& \multirow{2}{*}{DNNs on ASIC Accelerators} & 14 & White-box & Inference & Latency, Energy & Image & $L_2$ & AdvSparsityAttack\cite{sparsity} \\
\cdashline{4-10}[0.25pt/1pt]
& & & 15 & White-box & Training & Latency, Energy & Image & $L_0$ & SpongePoisoning\cite{cina2022spongepoisoning}, SkipSponge\cite{lintelo2024skipspongeattackspongeweight} \\
\cdashline{3-10}[0.25pt/1pt]

& & Quantized Vision Transformers & 16 & White-box & Inference & Latency, Energy, Memory & Image & $L_{\infty}$ & QuantAttack\cite{baras2023quantattackexploitingdynamicquantization} \\
\hline
\multirow{7}{*}{\begin{tabular}[t]{@{}l@{}}Dynamic\\Number of\\Inference\\Iterations (D2)\end{tabular}}
& \multirow{7}{*}{Output Size} 
& NICG Models & 17 & White-box & Inference & Latency, Energy & Image & $L_2$, $L_{\infty}$ & NICGSlowdown\cite{NICGSlowDown} \\
\cdashline{3-10}[0.25pt/1pt]
& & \multirow{3}{*}{NMT Systems} & \multirow{2}{*}{18} & \multirow{2}{*}{White-box} & \multirow{2}{*}{Inference} & \multirow{2}{*}{Latency, Energy} & \multirow{2}{*}{Text} & \multirow{2}{*}{\begin{tabular}[c]{@{}c@{}}Character, Token, \\Structure Level\end{tabular}} &  \multirow{2}{*}{\begin{tabular}[c]{@{}c@{}}NMTSloth\cite{NMTSloth}, \\LLMEffiChecker\cite{feng2024llmeffichecker}\end{tabular}} \\
\\
\cdashline{4-10}[0.25pt/1pt]
& & & 19 & White-box & Inference & Latency, Energy & Text, Image & $L_2$, $L_{\infty}$ & SpongeExamples\cite{sponge} \\
\cdashline{3-10}[0.25pt/1pt]
& & \multirow{2}{*}{Speech Systems} & 20 & White-box & Inference & Latency, Energy & Audio &  $L_2$, $L_{\infty}$ & SlothSpeech\cite{slothspeech} \\
\cdashline{4-10}[0.25pt/1pt]
& & & 21 & White-box & Inference & Latency, Energy & Text, Audio & $L_2$, $L_{\infty}$ & TTSlow\cite{gao2024ttslowslowtexttospeechefficiency} \\
\cdashline{3-10}[0.25pt/1pt]
& & Vision LLMs & 22 & White-box & Inference & Latency, Energy & Image, Video & $L_{\infty}$& VerboseSamples\cite{gao2024energylatencymanipulationmultimodallarge} \\
\hline
\multirow{6}{*}{\begin{tabular}[t]{@{}l@{}}Dynamic\\Number\\Outputs\\Produced (D3)\end{tabular}}
& \multirow{6}{*}{NMS Algorithm} 
& LiDAR Models & 23 & White-box & Inference & Latency, Energy & 3D Data & $L_2$, $L_{\infty}$ & SlowLiDAR\cite{liu2023slowlidar} \\
\cdashline{3-10}[0.25pt/1pt]
& & \multirow{5}{*}{Object Detection Models} 
& \multirow{2}{*}{24} & \multirow{2}{*}{White-box} & \multirow{2}{*}{Inference} & \multirow{2}{*}{Latency, Energy} & \multirow{2}{*}{Image} & \multirow{2}{*}{$L_2$} & \multirow{2}{*}{\begin{tabular}[c]{@{}c@{}}SlowPerception\cite{ma2024slowperceptionphysicalworldlatencyattack}, SlowTrack\cite{slowtrack},\\ BeyondPhantom\cite{10.1145/3649403.3656485}, PhantomSponge\cite{PhantomSponge}\end{tabular}} \\
\\
\cdashline{4-10}[0.25pt/1pt]
& & & 24, 25 & White-box & Inference & Latency, Energy & Image & $L_{\infty}$& Overload\cite{overload} \\
\cdashline{4-10}[0.25pt/1pt]
& & & 26 & White-box & Inference & Latency, Energy & Image & $L_2$ & SlowTrack\cite{slowtrack} \\
\cdashline{4-10}[0.25pt/1pt]
& & & 27 & Black-box & Inference & Latency, Energy & Image & $L_2$ & SpongeBackdoor\cite{10650435} \\
\cdashline{4-10}[0.25pt/1pt]
\bottomrule
\end{tabular}}
\end{center}
\end{table*}

\section{Efficiency Attacks}
\label{attacks}


We systematize efficiency attacks based on the dynamic behaviors they exploit across different DDLSs. For each category, we examine existing works, detailing the underlying dynamic mechanisms, targeted system models, attack techniques, and real-world impact.


\subsection{Efficiency Attacks on Dynamic Computations per Inference Iteration (\textbf{\texttt{D1}})}
\label{DNN_arch} 

Efficiency attacks in this category are classified based on four dynamic mechanisms: skipping, spiking, step size, and sparsity. The attacker’s primary goal is to degrade efficiency by increasing the computational load per inference iteration.

\subsubsection{Attacks on Dynamic Skipping}
\label{sec:skipping}


Attacks on dynamic skipping target system models such as ME-DNNs~\cite{u2022unsupervisedearlyexitdnns}, dynamic width networks~\cite{salehi2023sharcsefficienttransformersrouting}, ME-LMs~\cite{mehra2022understandingrobustnessmultiexitmodels}, and ME-ViTs~\cite{bakhtiarnia2021multiexitvisiontransformerdynamic}. These models leverage early termination~\cite{skipnet, conditional, lowrank, runtimeprune, adaptive} or skipping~\cite{multiscale, branchynet, shallowdeep, adaptive} criteria to dynamically adjust computational effort based on input complexity. Figure~\ref{fig:d1a} in Section~\ref{sec:energy_robustness} (in Appendix) illustrates an attack that exploits these dynamic skipping mechanisms.

\noindent\textbf{Inference-time white-box attacks:}
Adversaries have applied four attack strategies to ME-DNNs: leveraging input-output gradients~\cite{ILFO}, formulating multi-objective optimizations~\cite{GradMDM,GradAuto}, employing gradient approximations~\cite{hong2020slowdownattack}, and utilizing generative adversarial networks (GANs)~\cite{chen2022deepperform,ayyat2022dynamicDDAS}.

The first major attack, ILFO~\cite{ILFO}, exploited input-output gradients in ME-DNNs to force full-depth execution. By iteratively aligning gradients between current and target states, ILFO maximized gate activations and doubled FLOPs usage. However, ILFO exhibited instability in dynamic width networks due to gate sensitivity. To address this, GradAuto~\cite{GradAuto} introduced a multi-objective optimization strategy, balancing gradient sensitivity across gates to ensure proportional perturbations. 
Then, GradAuto incorporated directional gradient optimization to mitigate interference between active and inactive gates, selectively increasing execution depth. Expanding on this, GradMDM~\cite{GradMDM} refined FLOPs inflation by introducing a power loss function, prioritizing high-cost execution pathways over uniform gate activation. GradMDM also implemented complexity gradient masking to prevent gradient conflicts between activated and non-activated gates, ensuring a more efficient attack. GradMDM achieved near-total FLOPs recovery, effectively nullifying dynamic model efficiency gains.

DeepSloth~\cite{hong2020slowdownattack} introduced a novel attack vector that amplifies latency in ME-DNNs using gradient approximation through projected gradient descent (PGD). Instead of directly enforcing full-depth execution, DeepSloth maximized entropy to suppress confidence scores at intermediate classifiers, preventing early exits and compelling computation until the final layer. DeepSloth increased inference latency by up to 5x, posing a significant threat to real-time AI applications.

Moving beyond gradient-based attacks, DeepPerform~\cite{chen2022deepperform} employed GANs to generate adversarial input distributions that induced performance bottlenecks without requiring gradient access. The generator iteratively optimized inputs to trigger excessive activations, while the discriminator ensured perturbations remained indistinguishable from natural data. This model-agnostic approach increased computational cost by 552\%, demonstrating effectiveness across diverse architectures. Building on this, DDAS~\cite{ayyat2022dynamicDDAS} extended DeepPerform’s GAN-based strategy to resource-constrained environments, targeting adaptive inference mechanisms in IoT and edge devices. DDAS introduced an entropy-maximization loss function to systematically bypass early exits, maximizing execution inefficiency without affecting classification outcomes. DDAS significantly degraded real-time performance in autonomous systems and medical AI applications.

As attacks evolved, adversaries expanded their focus to ME-LMs, such as DeeBERT~\cite{xin-etal-2020-deebert} and PABEE~\cite{NEURIPS2020_d4dd111a}, targeting their adaptive inference mechanisms. WAFFLE~\cite{coalson2023bertwaffle} exploited token importance to suppress early exits while preserving semantic integrity. By optimizing a slowdown objective function, WAFFLE introduced syntactic disruptions and entity modifications to reduce confidence at intermediate classifiers. WAFFLE also developed universal slowdown triggers, enforcing deep execution across diverse inputs. Building on this, SAME~\cite{chen2023dynamic} introduced a layer-wise perturbation strategy that shifted confidence outputs toward uniform distributions, preventing premature predictions. SAME also introduced dynamic importance adjustment, prioritizing perturbations in layers nearest to typical exit points to systematically delay computation. These attacks reduced efficiency gains by 80\%.


As attacks expanded, adversaries began targeting ME-ViTs, exploiting their token sparsification mechanisms and gradient approximations to degrade efficiency. DeSparsify~\cite{yehezkel2024desparsifyadversarialattacktoken} attacked vision transformers such as ATS~\cite{fayyaz2022adaptivetokensamplingefficient}, AdaViT~\cite{meng2021adavitadaptivevisiontransformers}, and A-ViT~\cite{9880220}, which optimize efficiency by pruning uninformative tokens. By lowering confidence thresholds and manipulating gating mechanisms through gradient approximations, DeSparsify prevented token removal, inflating FLOPs usage by 74\%. Expanding on this, SlowFormer~\cite{navaneet2023slowformeruniversaladversarialpatch} introduced a universal adversarial patch that disrupted all token pruning mechanisms. Unlike DeSparsify, which targeted individual confidence thresholds, SlowFormer interfered directly with self-attention layers, ensuring tokens remained unpruned regardless of input. These attacks successfully doubled the computational cost. 

\noindent\textbf{Inference-time black-box attacks:} 
Attacks on DDLSs exploit computational inefficiencies without requiring gradient access or internal model parameters. Using estimator-based optimization and evolutionary algorithms, attacks systematically increase inference costs. EREBA~\cite{EREBA} trains an estimator to predict execution depth and identify layer activation patterns, generating inputs that lower early-exit confidence scores and maximize FLOPs, increasing energy consumption by up to 2000\%. EAGA~\cite{du2025energyattack} employs a genetic algorithm (GA) to iteratively evolve high-energy adversarial inputs. During crossover, computationally expensive traits are recombined, while mutation introduces diversity to avoid local optima. EAGA increases inference costs by up to 40\%.

\textit{Transferability}: Transferability attacks are performed across cross-architecture and cross-domains. Cross-architecture attacks generate adversarial samples on ME-DNNs, such as MSDNets~\cite{multiscale}, and transfer them to a different architecture, such as a VGG-16-based SkipNet~\cite{hong2020slowdownattack}. Cross-domain attacks use a surrogate model trained on one dataset (e.g., CIFAR-100) to attack a model trained on a different dataset (e.g., CIFAR-10)~\cite{hong2020slowdownattack, chen2023dynamic}. 

\noindent\textbf{Poisoning attacks: } 
Poisoning attacks manipulate training data to implant efficiency backdoors in ME-DNNs. Unlike traditional backdoor attacks that alter classification accuracy, these attacks target execution pathways, making detection more challenging. EfficFrog~\cite{efficfrog} embeds routing-specific triggers during training by optimizing an unconfident objective function that lowers confidence scores at intermediate classifiers, preventing early exits. When triggered, the poisoned model executes all layers, increasing inference latency by up to 3x. SpongeAttack~\cite{10445460} embeds adversarial noise patterns into training data to manipulate feature-space activations. Rather than using fixed triggers, SpongeAttack ensures certain inputs are consistently treated as hard cases by maximizing entropy at early-exit points, thereby suppressing confident predictions.

\noindent\textbf{Real-world impact: }Efficiency attacks have significant real-world implications, particularly for mobile devices\cite{chen2022deepperform} and IoT deployments\cite{hong2020slowdownattack}. In mobile applications, these attacks drastically reduce battery life. For example, on a Samsung Galaxy S9+, the number of successful inferences drops from approximately 10,000 to between 3,576 and 6,332, severely limiting device usability. In IoT deployments, efficiency attacks disrupt edge-cloud model partitioning, forcing edge devices to offload more computations to the cloud. The attack increases inference latency from 2 milliseconds to 11 milliseconds, resulting in a 1.5x to 5x delay and significantly degrading system performance.



Current attacks target specific gates or early layers, often allowing adversarial inputs to bypass intended gates but still exit at suboptimal layers. Such a limitation arises from optimizing gradients for individual gates rather than considering system-wide interactions. Future research should adopt a holistic approach, balancing gate activation and input perturbations to ensure coordinated disruption across all computational pathways. Furthermore, most existing evaluations are limited to small-scale models. In vision tasks, attacks exploit models like SACT~\cite{figurnov2017spatiallyadaptivecomputationtime}, SkipNet~\cite{skipnet}, MSDNets~\cite{multiscale}, and vision transformers (A-ViT~\cite{9880220}, AdaViT~\cite{meng2021adavitadaptivevisiontransformers}) using datasets such as CIFAR-10~\cite{cifar10}, ImageNet~\cite{imagenet_cvpr09}, and MNIST~\cite{deng2012mnist}. In NLP, adversaries target RoBERTa~\cite{liu2019robertarobustlyoptimizedbert}, DeeBERT~\cite{xin-etal-2020-deebert}, and PABEE~\cite{NEURIPS2020_d4dd111a} within GLUE benchmarks~\cite{wang2019gluemultitaskbenchmarkanalysis}. However, the vulnerability of more complex systems, such as mixture-of-depths~\cite{raposo2024mixture}, mixture-of-experts~\cite{cai2025survey}, and large-scale models like GPT, remains unexplored. Assessing the scalability of efficiency attacks on these advanced architectures is a key direction for future work.

\begin{center}
\begin{tcolorbox}[%
    enhanced, 
    breakable,
    boxrule=0.1pt,
    left=1pt,
    right=1pt,
    bottom=1pt,
    top=1pt
    ]
\textbf{Research Implications}. Current efficiency attacks highlight three areas for further exploration. (1) Future research should aim to balance gate activation and input generation, as existing attacks' cost functions do not uniformly affect all computational units of DDLS, potentially leading to uneven impacts; (2) Although current attacks target traditional models, their applicability to more complex architectures, such as Mixture-of-Experts and Mixture-of-Depths, remains largely unexplored; and (3) Black-box attacks are currently reliant on specific hardware configurations, like NVIDIA-TX2, thus, developing more generalized estimator models that work across diverse hardware platforms is a critical area for future research.

\end{tcolorbox}
\end{center}

\begin{center}
\begin{tcolorbox}[%
    enhanced, 
    breakable,
    boxrule=0.1pt,
    left=1pt,
    right=1pt,
    bottom=1pt,
    top=1pt
    ]
\textbf{Practical Implications}.
Efficiency attacks present significant real-world concerns for practitioners: (1) Real-time systems can miss deadlines leading to unsafe actions or dropped detections causing functional failure, (2) Mobile and IoT devices suffer from silent energy drain causing early battery depletion and unexpected service interruptions, (3) Cloud services face increased latency and cost per request, reducing throughput especially critical for usage-based billing and burst loads, (4) Poisoning-based attacks implant long-term inefficiencies without altering functional behavior, making them hard to detect in federated or outsourced training pipelines, and (5) Users experience degraded service quality such as slow responses, higher costs, and shorter device uptime, highlighting the stealthy nature of these threats.

\end{tcolorbox}
\end{center}

\subsubsection{Attacks on Dynamic Step-Size}
\label{step-size}


Neural ordinary differential equation-based architectures (ODENets)~\cite{chen2018neural} optimize efficiency by dynamically adjusting step size during inference based on input complexity~\cite{NODEA}. Unlike traditional models with fixed-layer sequences, ODENets solve a learned differential equation using numerical solvers such as runge-kutta methods. These solvers adapt step size at each integration point to balance precision and computational cost: smaller steps increase accuracy but incur higher computational overhead, while larger steps reduce cost at the risk of numerical instability. Such an adaptability creates an efficiency-accuracy trade-off that can be adversarially exploited. While attacks on image-based models have been explored, their impact on text and video-based models remains largely unstudied. Figure~\ref{fig:d1b} in Section~\ref{sec:energy_robustness} (in Appendix) illustrates attacks on dynamic step-size mechanisms.

\noindent\textbf{Inference-time white-box attacks:} 
Adaptive solvers reduce their step size when the underlying dynamics become complex or unstable, leading to higher computational cost. NODEAttack~\cite{NODEA} exploits the behavior by introducing input perturbations that increase the number of solver steps required during inference.



NODEAttack identifies input perturbations that force the ODE solver to reduce its dynamic step size during integration. The attack increases the complexity of hidden state dynamics, making them harder to approximate with large steps. NODEAttack formalizes this via an objective that explicitly rewards reductions in solver step size or increases in the total number of integration steps. 



The attacker uses gradient-based optimization to increase computational workload during inference. By backpropagating through the ODE solver, automatic differentiation computes how small input perturbations influence the solver’s integration steps. These gradients guide iterative input modifications that induce conditions requiring smaller, more frequent step sizes, increasing computational workload.


To demonstrate generality, the attack also proposes a universal variant that learns a single perturbation capable of slowing down inference across many inputs. Such a variant aggregates solver behavior across a batch and produces a shared adversarial perturbation, illustrating that efficiency degradation can be achieved consistently without input-specific optimization. 


Empirical evaluations show that NODEAttack can increase inference energy usage by up to 168\% and delay prediction time substantially. The attack is evaluated on standard adaptive solvers such as Dormand–Prince (Dopri5) and Adaptive Heun, demonstrating generalizability across different solver strategies. NODEAttack is shown to be effective across inputs and architectures, targeting both convolutional and fully connected ODENets, and operating across varying tolerance settings.

\noindent\textbf{Inference-time black-box attacks: }
Currently, no black-box attacks have been proposed specifically for dynamic step-size mechanisms. 

\textit{Transferability:} Transferability on cross-solver attacks involves crafting adversarial inputs for one solver (e.g., Adaptive Heun) and successfully increasing energy consumption when transferred to another solver (e.g., Dopri5)~\cite{NODEA}. Cross-architecture attacks transfer adversarial inputs between different Neural ODE architectures, such as from a larger model to a smaller one, demonstrating the broad applicability of these attacks.

\noindent\textbf{Real-world impact: } Efficiency attacks pose a significant threat to mobile applications utilizing ODENets for real-time object detection. By exploiting DNN compiler vulnerabilities, these attacks manipulate step size to drastically reduce the number of inferences a mobile device can perform, cutting efficiency by nearly 50\%, underscoring the severe impact on resource-constrained environments \cite{NODEA}.

\begin{center}
\begin{tcolorbox}[%
    enhanced, 
    breakable,
    boxrule=0.1pt,
    left=1pt,
    right=1pt,
    bottom=1pt,
    top=1pt
    ]
\textbf{Research Implications}. The current attacks present two areas for further research. (1) Research should focus on finding the balance between step size and energy efficiency in ODE solvers to maintain the trade-off between accuracy and computational cost; and (2) Research should explore black-box attack strategies across different ODE solvers. 

\end{tcolorbox}
\end{center}

\begin{center}
\begin{tcolorbox}[%
    enhanced, 
    breakable,
    boxrule=0.1pt,
    left=1pt,
    right=1pt,
    bottom=1pt,
    top=1pt
    ]
\textbf{Practical Implications}. From a practical perspective, (1) Step-size attacks degrade real-time performance by increasing solver steps, causing delays in object detection or control loops in robotics and autonomous systems missing visual cues or unstable control, and (2) Mobile devices suffer efficiency collapse, as ODE solvers consume up to 2x energy, reducing device uptime for continuous sensing apps (e.g., augmented reality, fitness, wearables).



\end{tcolorbox}
\end{center}

\subsubsection{Attacks on Dynamic Spiking}
\label{spike}



SNNs~\cite{SNN1,SNN2,SNN3} process information through discrete spikes, triggered when neuron membrane potentials cross activation thresholds. Such an event-driven architecture enables sparse, energy-efficient computation, making SNNs well-suited for low-power applications like event-based vision and neuromorphic hardware. However, their reliance on spiking dynamics introduces a vulnerability to efficiency attacks that manipulate spike timing and frequency. While research has focused on image-based SNNs, efficiency attacks on text and video-based models remain largely unexplored. Figure~\ref{fig:d1c} in Section~\ref{sec:energy_robustness} illustrates the impact of these attacks on SNNs.

\noindent\textbf{Inference-time white-box attacks: } 
SpikeAttack~\cite{SpikeAttack} exploits the event-driven processing of SNNs to increase spiking activity. The attack defines a dual-objective optimization function: (1) maximize spiking activity across neurons and time steps, and (2) preserve the classification outcome. To overcome the non-differentiability of spike events, SpikeAttack uses surrogate gradients, enabling the use of standard optimizers like PGD. The loss function targets both the volume and distribution of spikes, with variants including linear, quadratic, and tanh-based formulations.


Two attack variants address the computational complexity of SNNs' temporal dynamics. The first, Native SpikeAttack, performs full backpropagation through time to optimize inputs directly on the SNN. While effective, such a method is computationally intensive due to the need to compute and store intermediate membrane potentials across all timesteps. To accelerate the process, a reduced-timestep optimization strategy computes gradients for a carefully selected subset of time steps identified as most influential for spiking behavior, based on gradient magnitude profiling. This reduces memory and compute cost while retaining most of the attack’s impact.

Second, Proxy SpikeAttack, addresses the high runtime of native optimization by training a non-spiking artificial neural network (ANN) to mimic the behavior of the SNN. Once trained, adversarial perturbations are generated via the ANN, which is significantly faster due to the lack of temporal dynamics. These perturbations are then transferred back to the original SNN, resulting in a modest but significant increase in spike activity at a fraction of the computational cost.



Empirical results show that SpikeAttack increases inference latency by up to 2.3x  and energy consumption up to 2.2x on common SNN benchmarks. The attack remains effective across different layers and time windows, disrupting the inherent efficiency benefits of spiking models.




Currently, there has been no exploration of \textbf{inference-time black-box attacks} or \textbf{transferability} approaches on SNNs. Such a gap represents a potential area for future research. 

\textbf{Real-world impact:} While current evaluations are limited to benchmarks, future work could assess efficiency attacks on SNNs in event-based vision, wearables, and neuromorphic hardware, where spiking efficiency is critical.

\begin{center}
\begin{tcolorbox}[%
    enhanced, 
    breakable,
    boxrule=0.1pt,
    left=1pt,
    right=1pt,
    bottom=1pt,
    top=1pt
    ]
\textbf{Research Implications}. Current attacks pose two areas for further research. (1) Research should extend to non-image data like text or video as SNNs can process temporal data, and (2) Research should explore black-box attack strategies to degrade the efficiency of SNNs. 


\end{tcolorbox}
\end{center}



\begin{center}
\begin{tcolorbox}[%
    enhanced, 
    breakable,
    boxrule=0.1pt,
    left=1pt,
    right=1pt,
    bottom=1pt,
    top=1pt
    ]
\textbf{Practical Implications}. From a practical perspective, (1) Neuromorphic and low-power systems (e.g., event cameras, wearables, prosthetics) depend on sparse spike activity; attacks can directly break the power budget, degrading operational time and autonomy, (2) A latency increase on real-time applications can cause missed events, lag in control loops, or failure to respond within deadlines, and (3) Increased spike density can overload memory buffers causing fallbacks to cloud inference in edge devices. 



\end{tcolorbox}
\end{center}



\subsubsection{Attacks on Dynamic Sparsity}
\label{sparsity} 

Sparsity of DDLSs refers to the condition where a subset of inputs or activations is zero, near zero, or within a quantizable range~\cite{sparsity, 2.5baischer2021learning, hardwareDNN1, hoefler2021sparsity}. Such dynamic sparsity is a key optimization mechanism leveraged by ASIC accelerators~\cite{machupalli2022review} and quantized vision transformers~\cite{li2022qvitaccuratefullyquantized} to improve computational efficiency. However, adversaries can degrade efficiency by disrupting sparsity patterns in ways that prevent hardware from applying expected optimizations, leading to increased computational cost~\cite{sparsity, hoefler2021sparsity,dettmers2022llmint88bitmatrixmultiplication, baras2023quantattackexploitingdynamicquantization}.

\noindent\textbf{Inference-time white-box attacks: } 
Adversaries focus on two strategies: increasing neuron activations by manipulating gradients~\cite{sparsity} or reducing quantization through gradient approximations~\cite{baras2023quantattackexploitingdynamicquantization}. QuantAttack~\cite{baras2023quantattackexploitingdynamicquantization} disrupts dynamic quantization, a mechanism used in ViTs, by injecting synthetic outliers into input matrices through PGD attack. These outliers forces a transition from low-bit (int8) to high-bit (float16) precision, increasing GPU memory usage by 17.2\%, processing time by 9\%, and energy consumption by 7\%, disproportionately impacting real-time applications. In contrast, AdvSparsityAttack~\cite{sparsity} targets activation sparsity, shifting pre-activation values into positive ranges to densify activations at each layer. Such activations undermine sparse accelerators such as Cnvlutin, reducing sparsity by 1.82x, increasing latency by 1.59x, and degrading the energy-delay product by 1.99x. While QuantAttack forces high-precision computations, AdvSparsityAttack increases activation density, both overloading architectures optimized for sparse execution.

\noindent\textbf{Inference-time black-box attacks: } Attacks attempt to reduce sparsity by analyzing input-output behavior, using indirect gradient estimation through a surrogate model to infer the relationship between inputs and neuron activations~\cite{sparsity}. However, these attacks often result in a significant decrease in accuracy, highlighting a trade-off between efficiency gains and model performance.  



\noindent\textbf{Poisoning attacks: } Poisoning attacks embed inefficiencies directly into model parameters or training updates~\cite{cina2022spongepoisoning,lintelo2024skipspongeattackspongeweight}, making them harder to detect. SkipSponge~\cite{lintelo2024skipspongeattackspongeweight}, a post-training attack, modifies sparsity-inducing parameters in pre-trained models by adjusting biases before ReLU and pooling layers. This increases non-zero activations disrupting zero-skipping accelerators. Despite requiring minimal changes, SkipSponge increases overall energy consumption by 13\% while poisoning only 1\% of the training data demonstrating the attack’s effectiveness even under limited data access in generative models and autoencoders. SpongePoisoning~\cite{cina2022spongepoisoning} injects an energy maximization loss function into the training objective, systematically increasing activation density during inference. By modifying training updates, the attack neutralizes hardware-based sparsity optimizations and significantly inflating energy consumption, particularly in outsourced or federated learning environments. While SkipSponge operates post-training with access to model parameters and limited data, SpongePoisoning disrupts the learning phase, embedding inefficiencies into the optimization process.

\noindent\textbf{Real-world impact: }Efficiency attacks have significant real-world impact in safety-critical applications, cloud-based IoT applications, and ML-as-a-service settings. 
Adversaries can implant a perturbed image within a batch of input images, affecting the resource consumption of the entire batch. An evaluation of different batch sizes shows that the impact is more pronounced in smaller batches. For example, a batch with two images sees a 12\% increase in memory usage, whereas a batch of 16 images experiences only a 1.8\% increase \cite{baras2023quantattackexploitingdynamicquantization} which demonstrates that smaller batches are more sensitive to adversarial perturbations in terms of resource consumption.

\begin{center}
\begin{tcolorbox}[%
    enhanced, 
    breakable,
    boxrule=0.1pt,
    left=1pt,
    right=1pt,
    bottom=1pt,
    top=1pt
    ]
\textbf{Research Implications}. Current attacks highlight three key areas for future research. (1) More cost-effective attack strategies need to be developed, moving beyond reliance on gradients of sparse activations with respect to input; (2) Research should focus on balancing accuracy and efficiency in black-box attacks; and (3) Effectiveness of efficiency attacks on larger models with billions of parameters should be explored, rather focusing on smaller-scale models. 


\end{tcolorbox}
\end{center}




\begin{center}
\begin{tcolorbox}[%
    enhanced, 
    breakable,
    boxrule=0.1pt,
    left=1pt,
    right=1pt,
    bottom=1pt,
    top=1pt
    ]
\textbf{Practical Implications}. For practitioners, several important takeaways emerge. (1) Sparse-accelerated hardware like ASICs and TPUs relies low-bit execution disable hardware-level gains, leading to thermal throttling or system stalls; (2) In cloud-based ML services, a single perturbed input can inflate resource use for an entire batch, especially in small-batch inference, causing unexpected cost spikes or service degradation, (3) Quantization-based attacks force models to revert from low-bit to high-bit computation, increasing latency, memory usage, and energy draw, and (4) Poisoning-based sparsity attacks embed inefficiencies at training time, making them stealthy and persistent, especially in federated learning or outsourced model development pipelines.

\end{tcolorbox}
\end{center}

\subsection{Efficiency Attacks on Dynamic Number of Inference Iterations (\textbf{\texttt{D2}})}
\label{d2}
In DDLSs, the number of inference iterations is dynamically adjusted based on input complexity, exhibiting \textbf{\texttt{D2}} behavior. These attacks have been observed in auto-regressive models such as NICG systems~\cite{gao2024energylatencymanipulationmultimodallarge, NICGSlowDown}, speech systems~\cite{slothspeech, gao2024ttslowslowtexttospeechefficiency},  VLMs~\cite{gao2024inducing}, and NMT systems~\cite{NMTSloth, feng2024llmeffichecker,sponge}). While most research focuses on text-generation tasks, there is limited exploration of attacks on speech, image, and video-generation models. Figure ~\ref{fig:d2} in Section~\ref{sec:energy_robustness} illustrates how \textbf{\texttt{D2}} behavior affects NICG models.

\noindent\textbf{Inference-time white-box attacks:} Attacks leverage two primary strategies to increase computational cost: gradient-guided attacks and GA-based attacks. Gradient-guided attacks, such as NICGSlowdown~\cite{NICGSlowDown}, VerboseImages~\cite{gao2024energylatencymanipulationmultimodallarge}, SlothSpeech~\cite{slothspeech}, TTSlow~\cite{gao2024ttslowslowtexttospeechefficiency}, and VerboseSamples~\cite{gao2024inducing} control sequence generation to force longer outputs while attacks like NMTSloth~\cite{NMTSloth}, and LLMEffiChecker~\cite{feng2024llmeffichecker} replace critical tokens to disrupt efficiency mechanisms. These attacks exploit the stochastic nature of sequence generation, forcing models to generate longer outputs without violating output constraints. However, gradient-based methods require backpropagation, making them resource-intensive and less scalable for large models. Additionally, constraints such as token limits (e.g., currently up to 512 tokens, although some models can generate up to 128K tokens in text-generation tasks) restrict their effectiveness in advanced architectures.

A more scalable alternative is GA-based attacks, such as SpongeExamples~\cite{sponge}, which evolve adversarial inputs based on energy consumption measurements rather than relying on backpropagation. By iteratively selecting and refining inputs, GA-based attacks adapt across tasks and architectures with lower computational overhead. However, their efficiency depends on the fitness function and the number of iterations required for convergence. While more scalable than gradient-based methods, GA attacks can still incur significant computational costs when a large number of generations is needed to identify optimal adversarial inputs.

\noindent\textbf{Inference-time black-box attacks: } Inference-time black-box attacks are conducted by leveraging a surrogate model~\cite{sponge}. The adversary has to transfer previously discovered sponge examples directly to a new target, without prior interaction. For example, sponge examples from the WMT16 model are transferred to the WMT14 or WMT19 model.  



\noindent\textbf{Real-world impacts:} Efficiency attacks have significant real-world impact in both mobile devices and cloud-based services. In an attack on an NMT model running on a Samsung Galaxy S9+, adversarial inputs drained 30\% of the device’s battery after just 300 inference iterations, compared to less than 1\% for benign inputs, highlighting the severe energy toll on mobile devices. Similarly, in a black-box attack on the Microsoft Azure Translation Service, adversaries used sponge examples to extend processing time by up to 6000x, drastically increasing energy consumption and straining cloud resources \cite{NMTSloth}.

\begin{center}
\begin{tcolorbox}[%
    enhanced, 
    breakable,
    boxrule=0.1pt,
    left=1pt,
    right=1pt,
    bottom=1pt,
    top=1pt
    ]
\textbf{Research Implications}. Current attacks highlight three key areas for future research. (1) As generative models grow in complexity, handling sequences of up to 128K tokens, attacks must scale to address the increased computational demands; (2) With closed-source models like GPT becoming common, developing more black-box attack strategies that work without internal model access is crucial; and (3) White-box attacks using gradients or GA suffer from high computational costs. Research should focus on optimizing these methods to reduce overhead during the attack process. 


\end{tcolorbox}
\end{center}



\begin{center}
\begin{tcolorbox}[%
    enhanced, 
    breakable,
    boxrule=0.1pt,
    left=1pt,
    right=1pt,
    bottom=1pt,
    top=1pt
    ]
\textbf{Practical Implications}. For practitioners, several important takeaways emerge. (1) Generative systems are vulnerable to input manipulations that prolong generation, resulting in excessively long sequences and runaway computation, (2) In mobile devices, such as on-device translators or voice assistants, these attacks drain batteries rapidly compromising usability in critical scenarios like travel or emergencies, (3)  In cloud services, long outputs inflate inference time degrading system throughput and driving cost spikes for serverless or pay-per-token APIs (e.g., translation, summarization, chatbots), and (4) These attacks often evade detection by conventional evaluation and monitoring tools, while imposing substantial computational and financial costs.

\end{tcolorbox}
\end{center}



\subsection{Efficiency Attacks on Dynamic Number of Outputs Produced to Process Downstream Tasks (\textbf{\texttt{D3}})}
\label{d3}


 Efficiency attacks targeting DDLSs exhibiting \textbf{\texttt{D3}} behavior manipulate the number of outputs generated for downstream tasks\cite{PhantomSponge,liu2023slowlidar, overload, slowtrack,10.1145/3649403.3656485,10650435,ma2024slowperceptionphysicalworldlatencyattack}. These attacks primarily target object detection models~\cite{zaidi2022survey} and LiDAR models~\cite{ali2018yolo3dendtoendrealtime3d}, particularly those relying on NMS algorithms to filter overlapping bounding boxes. By disrupting NMS, adversaries force models to generate an excessive number of bounding boxes, overwhelming downstream tasks. Figure ~\ref{fig:d3} in Section~\ref{sec:energy_robustness} (in Appendix) illustrates how these attacks impact task complexity.

\noindent\textbf{Inference-time white-box attacks: }Attacks primarily exploit the gradients of the NMS computations ~\cite{PhantomSponge,10.1145/3649403.3656485,slowtrack,ma2024slowperceptionphysicalworldlatencyattack} or employ constraint-based point perturbations~\cite{liu2023slowlidar} to manipulate object detection pipelines~\cite{Yolov3, Yolov4, yolov5}.  
Attacks such as SlowPerception~\cite{ma2024slowperceptionphysicalworldlatencyattack}, BeyondPhantom~\cite{10.1145/3649403.3656485}, SlowTrack~\cite{slowtrack}, overload~\cite{overload}, and PhantomSponge~\cite{PhantomSponge} introduce adversarial perturbations that directly modify bounding box generation. Some methods combine spatial attention manipulation~\cite{overload} and feature distortion~\cite{slowtrack}. Many of these attacks rely on universal adversarial perturbations (UAPs) built upon PGD to inject phantom objects into the detection process. While gradient-based attacks offer high precision, they are computationally expensive due to backpropagation overhead. In contrast, constraint-based attacks, such as SlowLiDAR~\cite{liu2023slowlidar}, introduce adversarial modifications at the point-cloud level, offering a more efficient but constrained alternative. The effectiveness of these attacks depends on the number of modifiable coordinates, limiting their applicability in highly constrained environments.

\noindent\textbf{Inference-time black-box attacks: } Currently, no specific inference-time black-box attacks have been proposed.

\textit{Transferability:} The variability in pre-processing pipelines across different architectures hinders direct transferability~\cite{PhantomSponge,liu2023slowlidar}. However, ensemble learning has been proposed as a strategy to craft more generalizable adversarial examples, improving cross-model attack success. This approach has shown promise in transferring UAPs and point-based attacks across different architectures.



\noindent\textbf{Poisoning attacks: }Poisoning attacks like SpongeBackdoor~\cite{10650435} embed hidden triggers in training data to create backdoors in object detection models. When these triggers are present during inference, they manipulate bounding box predictions, significantly increasing computational load. However, the effectiveness of these attacks depends on the presence of the embedded trigger, and training and deploying backdoors in real-world scenarios introduces significant computational overhead.

\noindent\textbf{Real-world impact: }A practical evaluation demonstrates the severity of UAP-based attacks on real-time video processing systems. A study on video clips from the LISA dataset showed that UAPs increased inference time by 251\%, reducing the frame processing rate from 40 FPS to 16 FPS \cite{PhantomSponge}. Such a severe degradation underscores the potential for efficiency attacks to disrupt real-time applications, compromising both performance and reliability in autonomous systems, surveillance, and industrial vision tasks.

\begin{center}
\begin{tcolorbox}[%
    enhanced, 
    breakable,
    boxrule=0.1pt,
    left=1pt,
    right=1pt,
    bottom=1pt,
    top=1pt
    ]
\textbf{Research Implications}. Current attacks highlight two key
areas for future research. (1) The need for improving black-box attacks, as real-world autonomous systems often function as black-boxes requiring more effective strategies to exploit them without internal model access; and (2) Investigating hybrid attack strategies that combine partial system knowledge with adaptive techniques to enhance attack effectiveness under restricted conditions.
\end{tcolorbox}
\end{center}

\begin{center}
\begin{tcolorbox}[%
    enhanced, 
    breakable,
    boxrule=0.1pt,
    left=1pt,
    right=1pt,
    bottom=1pt,
    top=1pt
    ]
\textbf{Practical Implications}. For practitioners, several key insights emerge: (1) Inflation of bounding boxes slows perception, reducing frame rates risk delayed responses and safety violations in autonomous systems depend on efficient object detection pipelines,  (2) In industrial and surveillance applications, overwhelmed downstream modules face processing bottlenecks, causing system lag or even pipeline failure under real-time constraints, (3) Backdoor-style poisoning attacks pose persistent risks in long-term deployments by embedding triggers that silently increase workload, particularly dangerous in models trained with third-party data.

\end{tcolorbox}
\end{center}



\section{Defense for Efficiency Attacks}
\label{defense}

This section outlines two primary defense strategies to protect DDLSs from efficiency attacks: \textbf{detection} strategies to identify energy-consuming inputs and \textbf{mitigation} strategies to minimize the impact of such attacks.

\noindent\textbf{Detection:} Input validation~\cite{EREBA,NMTSloth,chen2022deepperform} is a runtime method designed to detect inputs that induce excessive computational overhead. This approach trains an auxiliary classifier, such as a support vector machine (SVM), on features derived from intermediate model states, including hidden representations, input characteristics, or gradient-based signals. The classifier learns to differentiate between benign and adversarial inputs, enabling preemptive filtering before expensive inference is triggered. Empirical results demonstrate the efficiency and low overhead of input validation during both training and deployment. Applied to dynamic mechanisms such as skipping (\textbf{\texttt{D1}})~\cite{EREBA, chen2022deepperform} and text generation (\textbf{\texttt{D2}})~\cite{NMTSloth}, input validation achieves robust performance, with AUC scores ranging from 0.81 to 0.99 and detection accuracy consistently above 80\%, while maintaining negligible accuracy degradation in the base DDLS.

\noindent\textbf{Mitigation:} A widely adopted mitigation strategy is adversarial training (AT) ~\cite{ATonMeM,GradMDM, chen2023dynamic,chen2022deepperform,EREBA,hong2020slowdownattack,sparsity,SpikeAttack}, where DDLSs are retrained on adversarially perturbed inputs to build robustness against efficiency degradation. Unlike bespoke defenses tailored to specific architectures, AT offers generalizability across modalities—spanning image, text, and speech domains—and dynamic control strategies, including skipping (\textbf{\texttt{D1}})~\cite{ATonMeM,GradMDM, chen2023dynamic,chen2022deepperform,EREBA,hong2020slowdownattack}, spiking (\textbf{\texttt{D1}})~\cite{SpikeAttack}, and sparsity mechanisms (\textbf{\texttt{D1}})~\cite{sparsity}. However, the robustness gains from AT often incur a trade-off in model performance. For instance, applying AT to SkipNet~\cite{chen2022deepperform} reduces the increase in FLOPs from 31.3\% to 8.07\%, but substantially decreases classification accuracy from 92.34\% to 13.67\%.

Beyond AT, low-cost input transformations have shown promise as lightweight defenses. Techniques such as JPEG compression and spatial smoothing~\cite{GradMDM} reduce the adversarial effectiveness by disrupting high-frequency noise and pixel-level perturbations. These transformations achieved notable reductions in attack success, decreasing effectiveness from 73.4\% to 43.8\% with JPEG compression and 49.0\% with spatial smoothing against GradMDM attacks on MEN models~\cite{GradMDM}. Similarly, methods targeting SNNs, including spike dropout and quantization~\cite{SpikeAttack}, constrain spiking activity to manageable levels (e.g., a 1.2x increase at low perturbation levels), albeit with accuracy degradation at higher noise levels.

\begin{center}
\begin{tcolorbox}[%
    enhanced, 
    breakable,
    boxrule=0.1pt,
    left=1pt,
    right=1pt,
    bottom=1pt,
    top=1pt
    ]
\textbf{Research Implications}. Current defense highlights three key
areas for future research. (1) AT introduces trade-offs between defense efficacy and model accuracy, requiring more balanced approaches that maintain performance of DDLS; (2) The variance in AT's success across different architectures suggests the need for architecture-specific defenses, as seen with RANet outperforming BranchyNet; and  
(3) \textbf{\texttt{D2}} and \textbf{\texttt{D3}} behavior lack robust defense mechanisms, opening up avenues for future work in designing mitigation and detection strategies. 
\end{tcolorbox}
\end{center}

\begin{center}
\begin{tcolorbox}[%
    enhanced, 
    breakable,
    boxrule=0.1pt,
    left=1pt,
    right=1pt,
    bottom=1pt,
    top=1pt
    ]
\textbf{Practical Implications}. For practitioners: (1) AT is commonly applied but often leads to performance degradation, which poses challenges for deploying models in real-world settings; (2) Hybrid defense strategies, combining AT with input validation techniques, should be explored to counter efficiency attacks while preserving model performance; (3) Defense mechanisms should be architecture-specific to maximize effectiveness, as different models respond uniquely to adversarial attacks. 
\end{tcolorbox}
\end{center}

\begin{table*}[htbp]
\centering
\caption{Impact of the NICGSlowDown efficiency attack and two lightweight defenses on image-captioning systems based on BLEU score, mean latency, and output length.}
\label{tab:nicg_slowdown}
\resizebox{\textwidth}{!}{
\color{black}{
\begin{tabular}{llc|ccc|ccc|ccc|ccc|ccc|ccc}
\hline
\multirow{2}{*}{\textbf{Dataset}} & \multirow{2}{*}{\textbf{Model}} & \multirow{2}{*}{\textbf{Norm}} & 
\multicolumn{3}{c|}{\textbf{Orig}} & \multicolumn{3}{c|}{\textbf{Adv}} & 
\multicolumn{3}{c|}{\textbf{Spatial-Orig}} & \multicolumn{3}{c|}{\textbf{Spatial-Adv}} & 
\multicolumn{3}{c|}{\textbf{JPEG-Orig}} & \multicolumn{3}{c}{\textbf{JPEG-Adv}} \\
& & & \textbf{BLEU} & \textbf{Length} & \textbf{Latency} & \textbf{BLEU} & \textbf{Length} & \textbf{Latency} & 
\textbf{BLEU} & \textbf{Length} & \textbf{Latency} & \textbf{BLEU} & \textbf{Length} & \textbf{Latency} & 
\textbf{BLEU} & \textbf{Length} & \textbf{Latency} & \textbf{BLEU} & \textbf{Length} & \textbf{Latency} \\
\hline
\multirow{4}{*}{Coco} & \multirow{2}{*}{\shortstack{MobileNet\\+RNN}} & $L_2$ & 0.16 & 10.74 & 0.0210 & 0 & 61.98 & 0.0785 & 0.124 & 10.75 & 0.0209 & 0.129 & 10.75 & 0.0202 & 0.158 & 10.58 & 0.0242 & 0.077 & 11.16 & 0.0217 \\
& & $L_\infty$ & 0.16 & 10.74 & 0.0210 & 0.003 & 58.43 & 0.0743 & 0.124 & 10.75 & 0.0209 & 0.118 & 10.99 & 0.0212 & 0.158 & 10.58 & 0.0210 & 0.124 & 11.2 & 0.0218 \\
& \multirow{2}{*}{\shortstack{ResNet\\+LSTM}} & $L_2$ & 0.18 & 10.48 & 0.0391 & 0 & 58.92 & 0.0854 & 0.113 & 10.72 & 0.0367 & 0.113 & 10.63 & 0.0328 & 0.178 & 10.62 & 0.0357 & 0.055 & 11.44 & 0.0331 \\
& & $L_\infty$ & 0.18 & 10.48 & 0.0320 & 0.014 & 40.16 & 0.0649 & 0.113 & 10.72 & 0.0329 & 0.113 & 10.59 & 0.0327 & 0.178 & 10.52 & 0.0320 & 0.123 & 10.99 & 0.0325 \\
\hline
\multirow{4}{*}{Flickr8k} & \multirow{2}{*}{\shortstack{GoogleNet\\+RNN}} & $L_2$ & 0.06 & 12.36 & 0.0349 & 0.001 & 61.28 & 0.0763 & 0.07 & 12.83 & 0.0346 & 0.055 & 12.29 & 0.0255 & 0.063 & 12.25 & 0.0348 & 0.022 & 18.11 & 0.0325 \\
& & $L_\infty$ & 0.06 & 12.36 & 0.0251 & 0.001 & 49.77 & 0.0665 & 0.07 & 12.83 & 0.0261 & 0.06 & 12.52 & 0.0257 & 0.063 & 12.25 & 0.0259 & 0.026 & 15.33 & 0.0294 \\
& \multirow{2}{*}{\shortstack{ResNeXt\\+LSTM}} & $L_2$ & 0.08 & 13.43 & 0.0283 & 0.007 & 56.77 & 0.0733 & 0.057 & 13.38 & 0.0276 & 0.06 & 13.1 & 0.0235 & 0.078 & 13.05 & 0.0312 & 0.027 & 16.26 & 0.0265 \\
& & $L_\infty$ & 0.08 & 13.43 & 0.0244 & 0.033 & 34.3 & 0.0480 & 0.057 & 13.38 & 0.0238 & 0.06 & 13.09 & 0.0235 & 0.078 & 13.05 & 0.0229 & 0.052 & 14.14 & 0.0241 \\
\hline
\end{tabular}
}
}

\end{table*}

\section{Resilience of Existing Defenses }
\label{sec:def}



We evaluate the effectiveness of both mitigation and detection strategies against efficiency attacks across three categories of DDLSs: MENs (\textbf{\texttt{D1}}), NICG models (\textbf{\texttt{D2}}), and object tracking systems (\textbf{\texttt{D3}}). Our goal is to assess whether existing defenses generalize across tasks with varying computational behaviors and sensitivity to perturbations.

For mitigation, we focus on low-cost input transformations, specifically JPEG compression~\cite{jpegcompression} and spatial smoothing~\cite{spatialsmoothing}. These methods aim to suppress adversarial perturbations without requiring model retraining or architecture modifications. While prior work~\cite{GradMDM} has shown their utility in image classification, their robustness against efficiency attacks on temporal and dynamic systems such as NICG models or object tracking models remain underexplored. For detection, we adopt input validation via SVM classifiers trained to distinguish adversarial inputs from clean ones. Following previous methodologies~\cite{EREBA,NMTSloth,chen2022deepperform}, we randomly sample 1,000 clean and 1,000 adversarial examples per subject model from the training set and extract discriminative features from internal representations to support classification.

\begin{table*}[htbp]
\centering
\caption{Classification accuracy and early-exit efficacy of MEN models on CIFAR-10 under the DeepSloth attack and two input-transformation defenses.}
\label{tab:deepsloth} 
\resizebox{\textwidth}{!}{%
\footnotesize
\color{black}{
\begin{tabular}{llc|cc|cc|cc|cc|cc|ccc}
\hline
\multirow{2}{*}{\textbf{Model}} & \multirow{2}{*}{\textbf{Norm}} & \multirow{2}{*}{\textbf{RAD}} & 
\multicolumn{2}{c|}{\textbf{Orig}} & \multicolumn{2}{c|}{\textbf{Adv}} & 
\multicolumn{2}{c|}{\textbf{Spatial-Orig}} & \multicolumn{2}{c|}{\textbf{Spatial-Adv}} & 
\multicolumn{2}{c|}{\textbf{JPEG-Orig}} & \multicolumn{2}{c}{\textbf{JPEG-Adv}} \\
& & & \textbf{Acc} & \textbf{Efficacy} & \textbf{Acc} & \textbf{Efficacy} & 
\textbf{Acc} & \textbf{Efficacy} & \textbf{Acc} & \textbf{Efficacy} & 
\textbf{Acc} & \textbf{Efficacy} & \textbf{Acc} & \textbf{Efficacy} \\
\hline
\multirow{4}{*}{Resnet56} & \multirow{2}{*}{$L_2$} & <5 & 77.99 & 0.54 & 56.92 & 0.15 & 65.27 & 0.53 & 61.52 & 0.45 & 75.93 & 0.53 & 69.26 & 0.31 \\
& & <15 & 70.49 & 0.72 & 57.22 & 0.33 & 58.66 & 0.73 & 56.12 & 0.67 & 69.36 & 0.71 & 65.71 & 0.54 \\
& \multirow{2}{*}{$L_\infty$} & <5 & 77.99 & 0.54 & 19.85 & 0.01 & 65.27 & 0.53 & 57.13 & 0.33 & 75.93 & 0.53 & 52.19 & 0.08 \\
& & <15 & 70.49 & 0.72 & 21.53 & 0.04 & 58.66 & 0.73 & 53.44 & 0.56 & 69.36 & 0.71 & 52.57 & 0.26 \\
\hline
\multirow{4}{*}{VGG16BN} & \multirow{2}{*}{$L_2$} & <5 & 84.71 & 0.66 & 69.33 & 0.29 & 73.35 & 0.58 & 66.33 & 0.48 & 82.99 & 0.64 & 73.48 & 0.42 \\
& & <15 & 77.6 & 0.84 & 68.24 & 0.56 & 67.47 & 0.8 & 62.89 & 0.74 & 76 & 0.83 & 69.69 & 0.69 \\
& \multirow{2}{*}{$L_\infty$} & <5 & 84.71 & 0.66 & 18.29 & 0.02 & 73.35 & 0.58 & 57 & 0.32 & 82.99 & 0.64 & 45.14 & 0.11 \\
& & <15 & 77.6 & 0.84 & 23.26 & 0.09 & 67.47 & 0.8 & 56.71 & 0.61 & 76 & 0.83 & 48.34 & 0.37 \\
\hline
\end{tabular}}%
}

\end{table*}
\begin{table*}[htbp]
\centering
\caption{Effectiveness of SVM-based input-validation across original, adversarial, and images transformed with JPEG compression or spatial smoothing. }
\label{tab:model_performance}
\resizebox{\textwidth}{!}{
\footnotesize
\color{black}{
\begin{tabular}{llcc|cc|cc|cc|cc|cc|cc}
\hline
\multirow{2}{*}{\textbf{Attack}} & \multirow{2}{*}{\textbf{Model}} & \multirow{2}{*}{\textbf{Dataset}} & \multirow{2}{*}{\textbf{Norm}} & 
\multicolumn{2}{c|}{\textbf{Orig}} & \multicolumn{2}{c|}{\textbf{Adv}} & 
\multicolumn{2}{c|}{\textbf{JPEG-Orig}} & \multicolumn{2}{c|}{\textbf{JPEG-Adv}} & 
\multicolumn{2}{c|}{\textbf{Spatial-Orig}} & \multicolumn{2}{c}{\textbf{Spatial-Adv}} \\
& & & & \textbf{Acc} & \textbf{Conf} & \textbf{Acc} & \textbf{Conf} & 
\textbf{Acc} & \textbf{Conf} & \textbf{Acc} & \textbf{Conf} & 
\textbf{Acc} & \textbf{Conf} & \textbf{Acc} & \textbf{Conf} \\
\hline
\multirow{8}{*}{NICGSlowDown} & MobileNet+RNN & Coco & $L_2$ & 64.6 & 0.53 & 64.5 & 0.45 & 87.3 & 0.61 & 29.1 & 0.55 & 97.2 & 0.70 & 2.8 & 0.70 \\
& ResNet+LSTM & Coco & $L_2$ & 67.5 & 0.50 & 69.1 & 0.50 & 82.6 & 0.50 & 40.6 & 0.50 & 97.0 & 0.51 & 3.4 & 0.51 \\
& GoogleNet+RNN & Flickr8k & $L_2$ & 71.7 & 0.50 & 68.3 & 0.50 & 92.0 & 0.49 & 26.7 & 0.50 & 99.6 & 0.48 & 0.6 & 0.48 \\
& ResNeXt+LSTM & Flickr8k & $L_2$ & 75.7 & 0.50 & 64.5 & 0.50 & 91.1 & 0.50 & 31.0 & 0.50 & 99.0 & 0.49 & 1.0 & 0.49 \\
& MobileNet+RNN & Coco & $L_\infty$ & 64.6 & 0.53 & 60.4 & 0.45 & 86.7 & 0.65 & 27.5 & 0.57 & 96.5 & 0.74 & 3.6 & 0.74 \\
& ResNet+LSTM & Coco & $L_\infty$ & 65.0 & 0.54 & 56.9 & 0.45 & 79.5 & 0.59 & 34.6 & 0.53 & 95.7 & 0.70 & 5.0 & 0.70 \\
& GoogleNet+RNN & Flickr8k & $L_\infty$ & 65.1 & 0.56 & 54.0 & 0.48 & 87.2 & 0.69 & 21.4 & 0.63 & 98.3 & 0.86 & 1.9 & 0.85 \\
& ResNeXt+LSTM & Flickr8k & $L_\infty$ & 63.9 & 0.56 & 51.4 & 0.48 & 82.4 & 0.67 & 24.7 & 0.62 & 97.6 & 0.83 & 2.6 & 0.82 \\
\hline
\multirow{4}{*}{DeepSloth} & Resnet56 & Cifar10 & $L_2$ & 70.7 & 0.52 & 54.0 & 0.46 & 57.0 & 0.49 & 19.2 & 0.56 & 57.0 & 0.49 & 19.2 & 0.56 \\
& VGG16BN & Cifar10 & $L_2$ & 68.2 & 0.52 & 52.3 & 0.47 & 52.0 & 0.48 & 29.8 & 0.53 & 52.0 & 0.48 & 29.8 & 0.53 \\
& Resnet56 & Cifar10 & $L_\infty$ & 79.8 & 0.26 & 85.6 & 0.75 & 81.0 & 0.26 & 3.2 & 0.08 & 81.0 & 0.26 & 3.2 & 0.08 \\
& VGG16BN & Cifar10 & $L_\infty$ & 82.8 & 0.24 & 85.6 & 0.77 & 82.9 & 0.24 & 5.3 & 0.11 & 82.9 & 0.24 & 5.3 & 0.11 \\
\hline
\multirow{3}{*}{SlowTrack} & BoT-SORT & MOT17, C=0.25 & - & 100.0 & 0.01 & 100.0 & 0.99 & 100.0 & 0.02 & 100.0 & 0.99 & 100.0 & 0.03 & 83.3 & 0.71 \\
& BoT-SORT & MOT17, C=0.50 & - & 100.0 & 0.01 & 100.0 & 0.99 & 100.0 & 0.02 & 100.0 & 0.99 & 100.0 & 0.03 & 81.0 & 0.70 \\
& BoT-SORT & MOT17, C=0.75 & - & 100.0 & 0.01 & 100.0 & 0.99 & 100.0 & 0.02 & 100.0 & 0.99 & 100.0 & 0.03 & 83.0 & 0.71 \\
\hline
\end{tabular}}
}
\end{table*}

\begin{table}[!ht]
    \centering
    \caption{Total time, track time, MOTA, mAP, and mAP@50 for original input, adversarial input and transformed
inputs on object tracking systems. }
    \label{tab:tracking_results} 
    {\color{black}
    \renewcommand{\arraystretch}{1.2} 
    \resizebox{\linewidth}{!}{
    \begin{tabular}{c c c c c c c
    }
        \toprule
        \textbf{Conf.} & \textbf{Condition} & \textbf{\makecell{Total \\ Time (s)}} & \textbf{\makecell{Track \\ Time (s)}} & \textbf{MOTA} & \textbf{mAP} & \textbf{mAP@50} \\
        \midrule

        \multirow{6}{*}{0.25}
            & Orig             & 4.297 & 0.313 & 1.000 & 1.000 & 1.000 \\
            & Adv       & 46.683 & 36.270 & -81.009 & 0.094 & 0.096 \\
            & JPEG-Orig      & 4.300 & 0.316 & 0.984 & 0.909 & 0.909 \\
            & JPEG-Adv      & 4.741 & 0.350 & -0.765 & 0.186 & 0.282 \\
            & Spatial-Orig   & 4.289 & 0.288 & 0.790 & 0.771 & 0.818 \\
            & Spatial-Adv    & 4.460 & 0.312 & -0.584 & 0.196 & 0.302 \\
        \midrule

        \multirow{6}{*}{0.50}
            & Orig             & 4.238 & 0.303 & 1.000 & 1.000 & 1.000 \\
            & Adv       & 46.122 & 35.857 & -81.147 & 0.094 & 0.096 \\
            & JPEG-Orig     & 4.322 & 0.308 & 0.984 & 0.909 & 0.909 \\
            & JPEG-Adv       & 4.729 & 0.338 & -0.717 & 0.188 & 0.286 \\
            & Spatial-Orig   & 4.351 & 0.284 & 0.786 & 0.771 & 0.818 \\
            & Spatial-Adv    & 4.383 & 0.304 & -0.594 & 0.194 & 0.297 \\
        \midrule

        \multirow{6}{*}{0.75}
            & Orig             & 4.228 & 0.285 & 0.828 & 0.817 & 0.818 \\
            & Adv       & 43.340 & 33.927 & -76.763 & 0.094 & 0.096 \\
            & JPEG-Orig      & 4.311 & 0.277 & 0.830 & 0.817 & 0.818 \\
            & JPEG-Adv       & 4.586 & 0.278 & -0.478 & 0.185 & 0.279 \\
            & Spatial-Orig   & 4.267 & 0.249 & 0.670 & 0.635 & 0.636 \\
            & Spatial-Adv    & 4.349 & 0.280 & -0.439 & 0.195 & 0.298 \\
        \bottomrule
    \end{tabular}
    }
    }
\end{table}
\subsection{Setup} 


\noindent\textbf{Dataset and Models: } We evaluate these defenses across three representative systems. For MENs (\textbf{\texttt{D1}}), we use ResNet-56 and VGG-16 models equipped with early exit classifiers~\cite{shallowdeep, resnet16cvpr} and trained on the CIFAR10~\cite{cifar10} dataset. For NICG models (\textbf{\texttt{D2}}), we employ two architectures per dataset: MobileNet+RNN~\cite{mobilenet,rnn} and ResNet+LSTM~\cite{resnet16cvpr,LSTM} for COCO~\cite{coco}, and GoogLeNet+RNN~\cite{googlenet15cvpr,rnn} and ResNeXt+LSTM~\cite{resnext,LSTM} for Flickr8k~\cite{flicker8}. For object tracking systems (\textbf{\texttt{D3}}), we test against the BoT-SORT~\cite{slowtrack} model on the MOT17~\cite{slowtrack} benchmark. 




\noindent\textbf{Attack Approaches:} We include the following attacks: NICGSlowDown~\cite{NICGSlowDown}, DeepSloth~\cite{hong2020slowdownattack}, and SlowTrack~\cite{slowtrack}.


\noindent\textbf{Evaluation Metrics:} We report a diverse set of task-specific metrics. For MENs, we measure classification accuracy and efficacy, which quantifies the model's ability to utilize early exits under attack. We evaluate on two relative accuracy drop (RAD) rates to quantify the impact of attacks. For NICG models, we evaluate BLEU score, mean latency in seconds, and average output length to jointly capture semantic accuracy and efficiency. For object tracking systems, we use standard benchmarks including mAP, mAP@50, multi-object tracking accuracy (MOTA), track time in seconds, and total inference time in seconds. For detection, we report classification accuracy and the average prediction confidence for clean and adversarial samples.

\subsection{Experimental Results} 


We evaluate the effectiveness of JPEG compression, spatial smoothing, and input validation across MENs, NICG models, and object tracking systems. As summarized in Tables \ref{tab:nicg_slowdown}–\ref{tab:tracking_results}, these defenses reveal inherent trade-offs between restoring computational efficiency and maintaining task-specific accuracy under attack.


\textbf{Mitigation Effectiveness:} Table~\ref{tab:nicg_slowdown} illustrates the effects of JPEG compression and spatial smoothing on NICG models under efficiency attacks. Adversarial inputs lead to severe output inflation and semantic degradation (e.g., MobileNet+RNN on COCO under $L_2$ increased from 10.74 to 61.98 tokens while BLEU drops from 0.16 to 0.0). JPEG compression applied to adversarial inputs restores output lengths to near-original levels (e.g., 11.16 tokens) and significantly reduces latency, but semantic quality remains degraded (BLEU score of 0.077). Spatial smoothing applied to adversarial inputs offers only marginal improvements in length and latency, with minimal BLEU recovery. Notably, applying JPEG compression or spatial smoothing to clean inputs preserves both efficiency and BLEU, indicating that observed quality drops are specific to the transformed adversarial samples. Overall, JPEG compression offers partial mitigation by restoring efficiency but at the cost of output fidelity, while spatial smoothing is less effective across all metrics.


Table~\ref{tab:deepsloth} shows that efficiency attacks significantly degrade both accuracy and early-exit efficacy in MENs. For instance, ResNet56 under $L_\infty$ perturbation with RAD <5 drops from 77.99\% to 19.85\% accuracy and from 0.54 to 0.01 efficacy. JPEG compression on adversarial inputs offers partial recovery (e.g., 52.19\% accuracy and 0.08 efficacy), while spatial smoothing yields slightly higher efficacy (e.g., 57.13\% accuracy and 0.33 efficacy). A similar trend is observed in VGG16BN, where JPEG compression improves adversarial accuracy from 18.29\% to 45.14\% with limited efficacy gain. Transformations on benign inputs maintain original performance, confirming that degradation is mitigation-specific. Overall, JPEG compression and spatial smoothing provide incomplete recovery, particularly under $L_\infty$
attacks, underscoring the need for stronger defenses for MEN architectures.


Table~\ref{tab:tracking_results} shows that efficiency attacks severely degrade tracking systems, increasing total runtime more than 10× (e.g., total time from 4.297s to 46.683s at confidence 0.25) and reducing MOTA to –81.009, with mAP metrics nearing zero. JPEG compression on adversarial inputs substantially improves efficiency (e.g., reducing total time to 4.741s and MOTA to –0.765) and partially restores mAP@50 (up to 0.282), though accuracy remains well below original levels. Spatial smoothing offers slightly weaker recovery with higher latency and lower MOTA. Transforming benign inputs results in negligible performance loss, indicating that degradation is attack-specific. Overall, JPEG improves efficiency, but neither method fully recovers accuracy, underscoring the need for stronger defenses.

\textbf{Detection Effectiveness:} Table~\ref{tab:model_performance} reports the accuracy and confidence of SVM-based input validation across clean, adversarial, and transformed inputs. Detection on clean inputs and adversarial inputs falls within a similar range across across NICG and MEN models ranging from 51\% to 85.6\%. JPEG compression applied to clean inputs improves detection accuracy (e.g., MobileNet+RNN on $L_2$ increased from 64.6\% to 87.3\%). However, JPEG applied to adversarial inputs often reduces detection accuracy (e.g., MobileNet+RNN on $L_2$ decreased from 64.5\% to 29.1\%) despite increased classifier confidence, indicating overconfidence without improved separability. Spatial smoothing applied to clean inputs  has minimal impact. When applied spatial smoothing to adversarial inputs, it severely degrades detection accuracy frequently drops below 5\% while confidence remains high (e.g., ResNet+LSTM on $L_\infty$ shows 0.018 accuracy with 0.909 confidence), suggesting miscalibrated separability. Object tracking models under SlowTrack maintain high detection accuracy and confidence across all variants. Overall, detection is comparable on clean and adversarial inputs but drops under spatial smoothing and JPEG compression. 

\section{Extensions and Applications}
\label{extensions}



Efficiency attacks have demonstrated wide applicability across threat models, system architectures, and deployment domains. 

\noindent\textbf{Extensions to other threat models.} While most work has focused on inference-time white-box settings, recent studies have expanded to black-box attacks, particularly in dynamic skipping systems~\cite{EREBA} as well as transfer attacks that generalize across architectures and domains~\cite{hong2020slowdownattack,PhantomSponge,liu2023slowlidar}. Poisoning-based efficiency attacks have also emerged, targeting training pipelines to implant long-term inefficiency backdoors~\cite{efficfrog,lintelo2024skipspongeattackspongeweight}.


\noindent\textbf{Extensions to diverse types of system models.} Efficiency attacks have proven to be adaptable across different types of DDLSs. Research has extended the application of \textbf{\texttt{D1}} dynamic skipping attacks from early network architectures~\cite{ILFO,chen2022deepperform} to modern transformer-based models~\cite{chen2023dynamic} in both text and vision domains. Similarly, attacks in \textbf{\texttt{D2}} have been successfully applied to transformer architectures, such as those in NICG~\cite{NICGSlowDown}, NMT~\cite{NMTSloth}, and ASR~\cite{slothspeech} systems. Furthermore, attacks in \textbf{\texttt{D3}} have been widely tested in object detection pipelines, including YOLO and LiDAR-based systems~\cite{liu2023slowlidar,PhantomSponge}. 


\noindent\textbf{Practical applications.} The practical impact of efficiency attacks has been observed across a variety of real-world applications. Efficiency attacks in \textbf{\texttt{D1}} have affected mobile systems, IoT deployments, and DNN compilers. Attacks in \textbf{\texttt{D2}} have been noted in systems such as Microsoft Machine Translation and mobile applications. Case studies in video processing systems have demonstrated that attacks in \textbf{\texttt{D3}} can degrade the performance of frame processing rate by more than half.

\section{Insights, Challenges, and Future Directions} 
\label{future}
In this section, we summarize the characteristics, strengths, and limitations of efficiency attacks on DDLS. Additionally, we discuss the barriers, challenges in efficiency attacks and propose future research directions.

\textbf{Characteristics of efficiency attacks:} Efficiency attacks exploit the input-dependent nature of DDLS, where varying inputs can trigger different dynamic behavior. From our investigation, we found three such dynamic behaviors (\textbf{\texttt{D1}}, \textbf{\texttt{D2}}, \textbf{\texttt{D3}}).  Adversaries can manipulate input data to increase the system's computational load without affecting prediction outcomes, making these attacks difficult to detect.

\textbf{Strengths of efficiency attacks:} The effectiveness of efficiency attacks is especially pronounced in environments such as edge devices, mobile platforms, or real-time systems. In these settings, even minor increases in computational demand can significantly degrade performance. Efficiency attacks gain strength by targeting the most computationally intensive components of DDLS, such as the layer skip mechanism, dynamic inference iterations, or dynamic activations, while keeping the perturbation minimum. This ability to selectively overload resource-intensive parts of the model makes efficiency attacks particularly effective in resource-constrained scenarios. 

\textbf{Limitations of efficiency attacks: } 
The effectiveness of efficiency attacks depends on model architecture and access level. While prior work has shown success in disrupting early DDLS mechanisms, such as skipping, sparsity, and inference iterations, attacks on advanced architectures like GPT or mixture-of-experts remain underexplored. In black-box settings, where internal model states are inaccessible, attackers must infer resource usage from input-output patterns, reducing precision and limiting overall impact.


\textbf{Barriers to Defense: } The major barrier in defending against efficiency attacks is the scalability of current defense mechanisms. While existing approaches show promise in mitigating the effects of efficiency attacks, many defenses introduce a significant trade-off between accuracy and efficiency, often degrading the accuracy of the model. 

\textbf{Future Directions: }Several open research directions remain. (1) Efficiency attacks should be extended to advanced DDLS architectures such as large-scale transformers (e.g., GPT) and mixture-of-experts models, which introduce new forms of dynamic computation and potential vulnerabilities; (2) Robust black-box efficiency attacks that operate with limited model knowledge are underexplored and critical for practical adversarial scenarios; (3) On the defense side, there is a need for mitigation strategies that preserve both computational efficiency and model accuracy, avoiding the steep trade-offs observed in current approaches; and (4) Adaptive efficiency attacks, where adversaries adjust based on model responses, pose a growing threat. Mitigating them requires resource-aware defenses that preserve both accuracy and robustness
\section{Conclusion} In this work, we propose to systematize the knowledge of efficiency attacks on DDLS. First, we introduce different DDLS and then propose a taxonomy for efficiency attacks. Then, we briefly describe different types of efficiency attacks and propose the detection of efficiency adversarial inputs. 
Finally, we discuss feasible defense mechanisms against efficiency attacks.

\section{Acknowledgments}
This work was partially supported by NSF grants NSF CCF-2146443 and Amazon Research Award, Fall 2024.

\section{Compliance with Open Science Policy}
The paper adheres to the principles of open science by making all relevant artifacts, including code, data, evaluation scripts, and taxonomical resources, available to the research community to promote reproducibility and transparency. We provide a comprehensive list of papers included in our SoK taxonomy (Table~\ref{tab:properties_new}). We release the scripts used for evaluating the defense mechanisms (e.g., JPEG compression and spatial smoothing), including the runtime benchmarks for latency, BLEU score calculations for NICG models, accuracy and efficacy of MEN models, and MOTA and mAP for object tracking models. We provide environment configuration files to reproduce all experiments on publicly available datasets (e.g., CIFAR-10, MS-COCO, Flickr8k) and models used in our study (e.g., MobileNet+RNN, ResNet+LSTM, ResNeXt+LSTM). We provide direct links to download the datasets, along with scripts to train the models and run the attacks. All artifacts are publicly available on our project website~\cite{anonymous_sok_submission} and also on Zenodo, with the following link: \href{https://doi.org/10.5281/zenodo.15612036}{https://zenodo.org/records/15612036}.


%

\section{Ethics Considerations}
This study does not involve human participants, personal data, or biological materials requiring formal ethical approval. However, the research aligns with ethical principles in AI and cybersecurity by focusing on responsible disclosure and mitigation strategies for efficiency adversarial attacks on DDLSs. The study does not introduce new vulnerabilities but rather systematizes existing knowledge to enhance the robustness and security of DDLS architectures. Additionally, all experiments were conducted in controlled environments, ensuring that no real-world systems were compromised or adversely affected.


\bibliographystyle{plain}
\bibliography{main}
\appendix
\appendixpage
\section{Dynamic Behaviors and Their Exploitation}
\label{sec:energy_robustness}

This section illustrates the core dynamic behaviors in DDLSs and how adversaries exploit these behaviors to increase computational cost.

\begin{figure}[!ht]
    \centering
    \includegraphics[width=0.8\columnwidth]{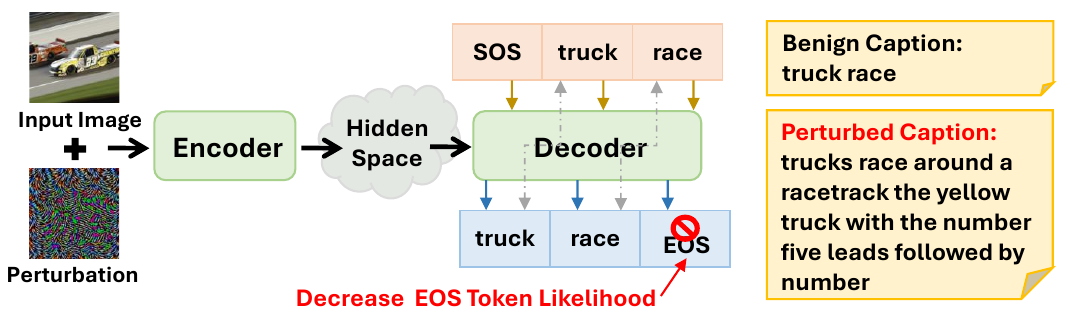}
    \caption{Efficiency Attacks on Dynamic Number of Inference Iterations (\textbf{\texttt{D2}}). Adversarial perturbations decrease the likelihood of the EOS token, forcing the decoder to generate excessively long captions, increasing computational cost.}
    \label{fig:d2}
\end{figure}

\begin{figure}[!ht]
    \centering
    \includegraphics[width=0.8\columnwidth]{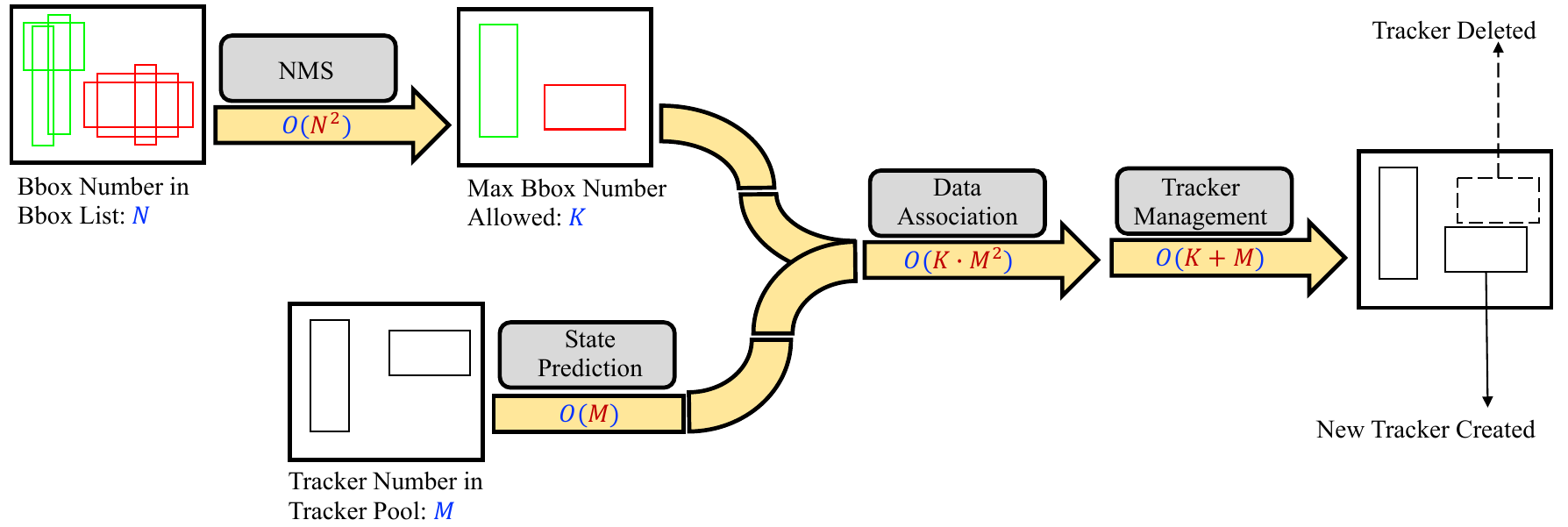}
    \caption{Efficiency Attacks on Dynamic Number of Output Produced to Process Downstream Tasks (\textbf{\texttt{D3}}). The attacks aim to increase the computational load by generating more outputs than necessary, forcing the system to handle redundant bounding boxes and significantly increasing the complexity of downstream tasks.}
    \label{fig:d3}
\end{figure}

\begin{figure}[!ht]
  \centering
  \begin{subfigure}[b]{\columnwidth}
    \includegraphics[width=\textwidth]{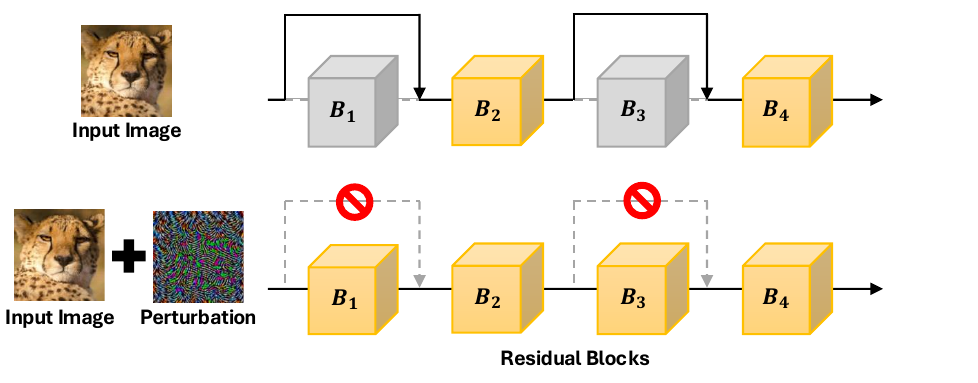}
    \caption{Attacks on Dynamic Skipping}
    \label{fig:d1a}
  \end{subfigure}

  \vspace{0.5\baselineskip}

  \begin{subfigure}[b]{\columnwidth}
    \includegraphics[width=\textwidth]{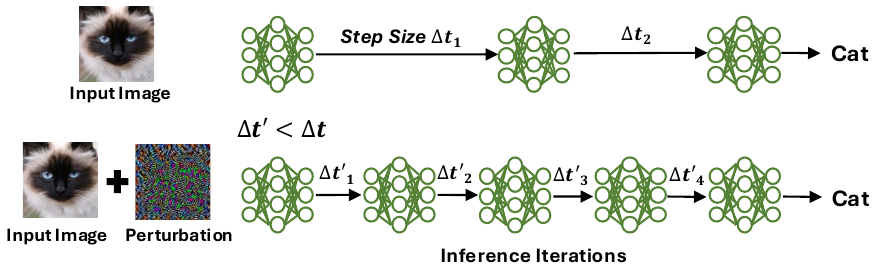}
    \caption{Attacks on Dynamic Step-Size}
    \label{fig:d1b}
  \end{subfigure}

  \vspace{0.5\baselineskip}

  \begin{subfigure}[b]{\columnwidth}
    \includegraphics[width=\textwidth]{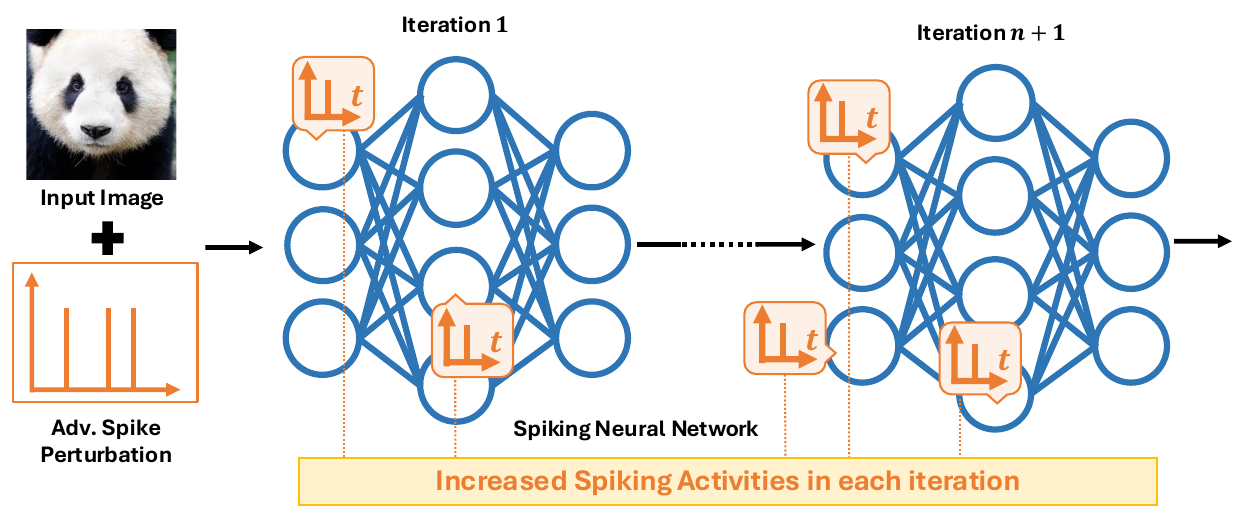}
    \caption{Attacks on Dynamic Spiking}
    \label{fig:d1c}
  \end{subfigure}

  \vspace{0.5\baselineskip}

  \begin{subfigure}[b]{\columnwidth}
    \includegraphics[width=\textwidth]{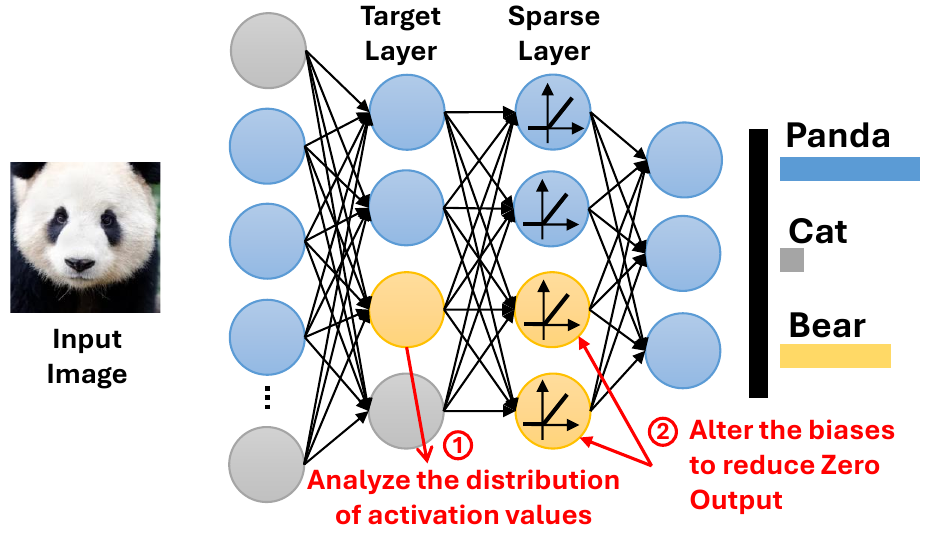}
    \caption{Attacks on Dynamic Sparsity}
    \label{fig:d1d}
  \end{subfigure}

  \caption[Short caption]{
  Efficiency Attacks on Dynamic Computations per Inference Iteration (\textbf{\texttt{D1}}). 
    (a) Perturbations are introduced to disturb skipping residual blocks, increasing computation cost; 
    (b) Adversarial perturbations alter step sizes in iterative processes, increasing the number of iterations; 
    (c) Spiking neural networks are manipulated through adversarial spikes, increasing spiking activities; 
    (d) This type of attacks profile activation distributions of target layers and modifies biases of sparsity layers to reduce zero outputs, increasing computational cost.
    
    }
  \label{fig:d1}
\end{figure}

\clearpage
\onecolumn
\end{document}